\documentclass[10pt,twocolumn,letterpaper]{article}

\usepackage{iccv}
\usepackage{times}
\usepackage{epsfig}
\usepackage{graphicx}
\usepackage{amsmath}
\usepackage{amssymb}
\usepackage{easyReview}
\usepackage{colortbl}
\usepackage{xcolor}
\usepackage{tikz}
\usepackage{subcaption}

\usepackage{hhline}
\usepackage{makecell}
\usepackage{array}

\def\checkmark{\tikz\fill[scale=0.4](0,.35) -- (.25,0) -- (1,.7) -- (.25,.15) -- cycle;} 

\definecolor{light-gray}{RGB}{240, 240, 240}

\DeclareMathOperator*{\argmax}{arg\,max}

\usepackage[pagebackref=true,breaklinks=true,letterpaper=true,colorlinks,bookmarks=false]{hyperref}

\iccvfinalcopy 

\def\iccvPaperID{35} 

\ificcvfinal\fi

\begin{document}

\newcommand{\figcifarer}[3]{
\begin{figure}[tb!]
\centering
\includegraphics[trim=#1, clip, width=#2\linewidth]{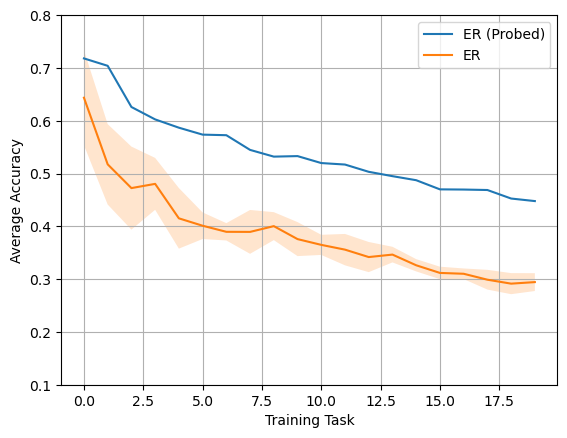}
\caption{#3}
\label{fig:probing_cifar100}
\end{figure}
}

\newcommand{\figcifarcomp}[3]{
\begin{figure*}[tb!]
\centering
\includegraphics[trim=#1, clip, width=#2\linewidth]{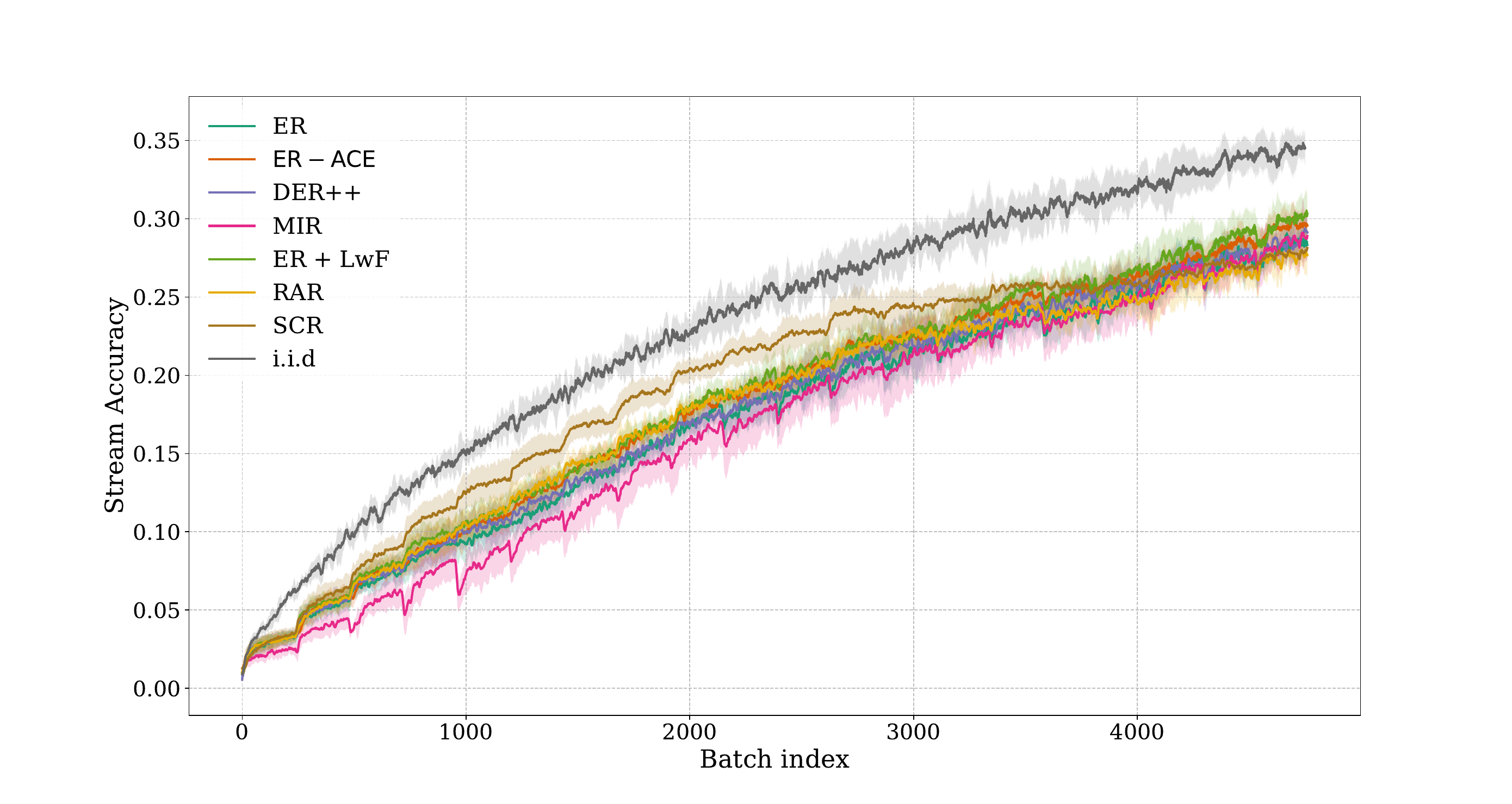}
\caption{#3}
\label{fig:comparison_cifar100}
\end{figure*}
}

\newcommand{\figtinycomp}[3]{
\begin{figure*}[tb!]
\centering
\includegraphics[trim=#1, clip, width=#2\linewidth]{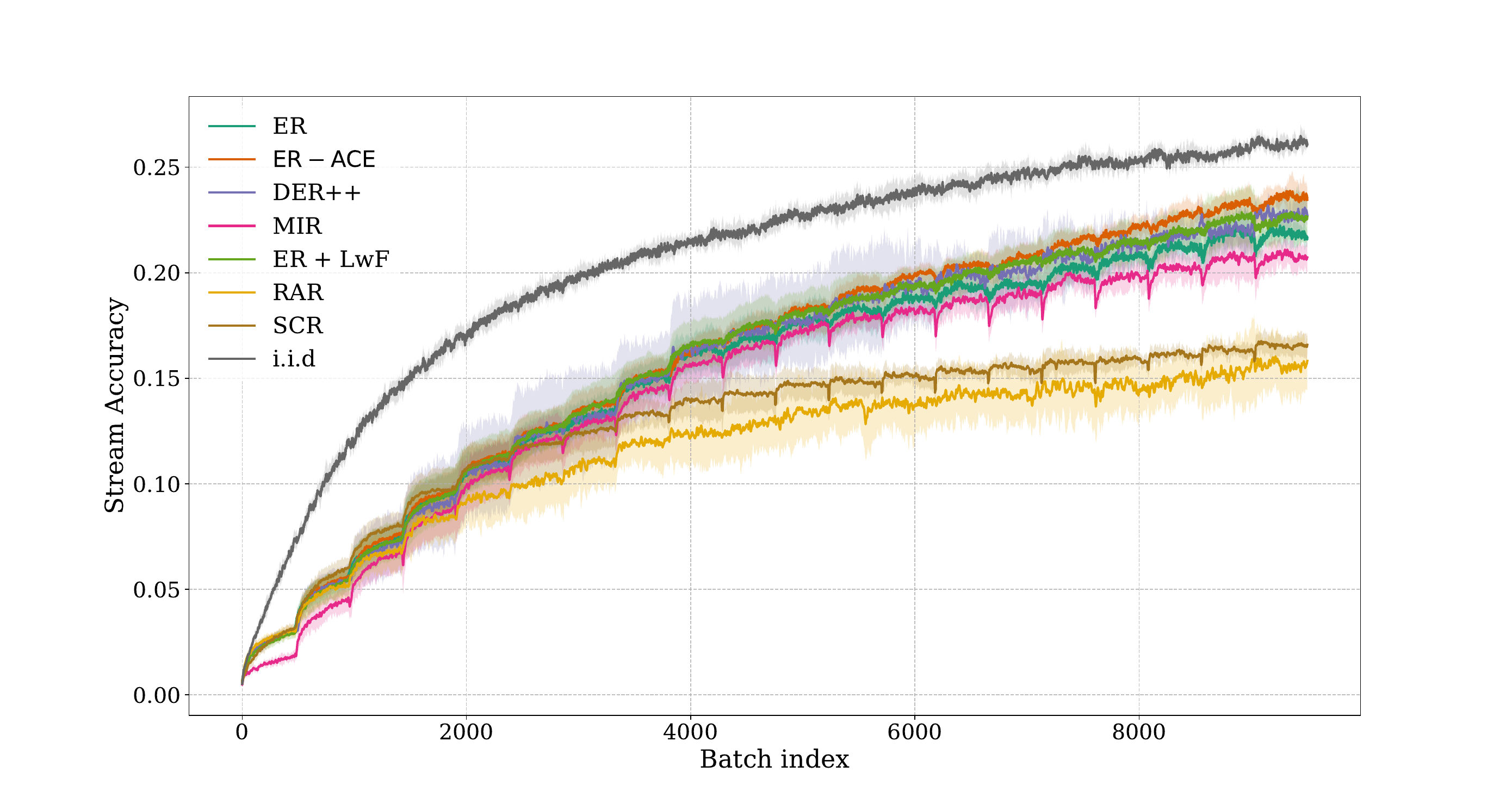}
\caption{#3}
\label{fig:comparison_tiny}
\end{figure*}
}

\newcommand{\figcomparison}[3]{
\begin{figure*}[tb!]
\centering
\includegraphics[trim=#1, clip, width=#2\linewidth]{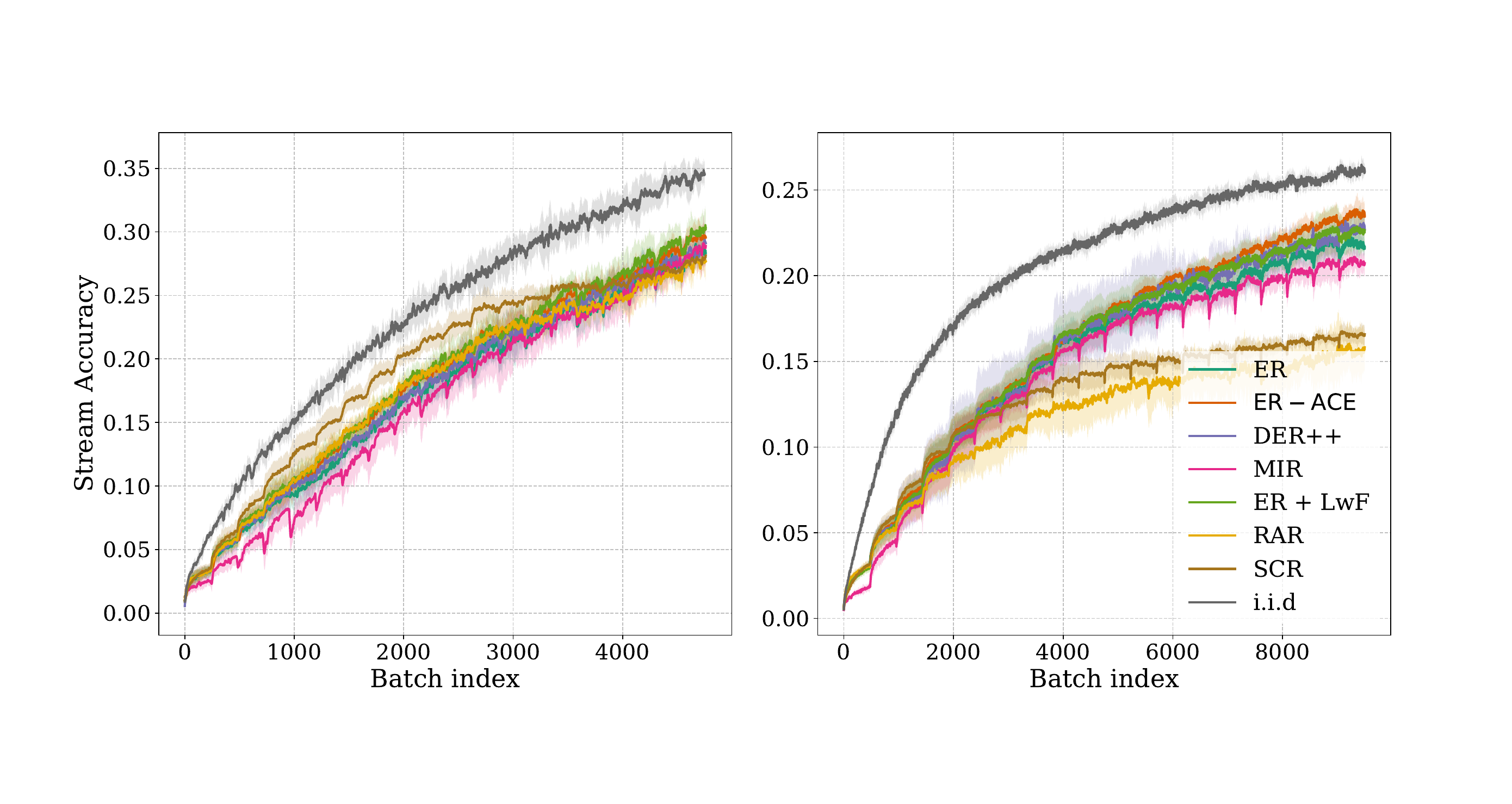}
\caption{#3}
\label{fig:comparison}
\end{figure*}
}

\newcommand{\figcifarforg}[3]{
\begin{figure}[tb!]
\centering
\includegraphics[trim=#1, clip, width=#2\linewidth]{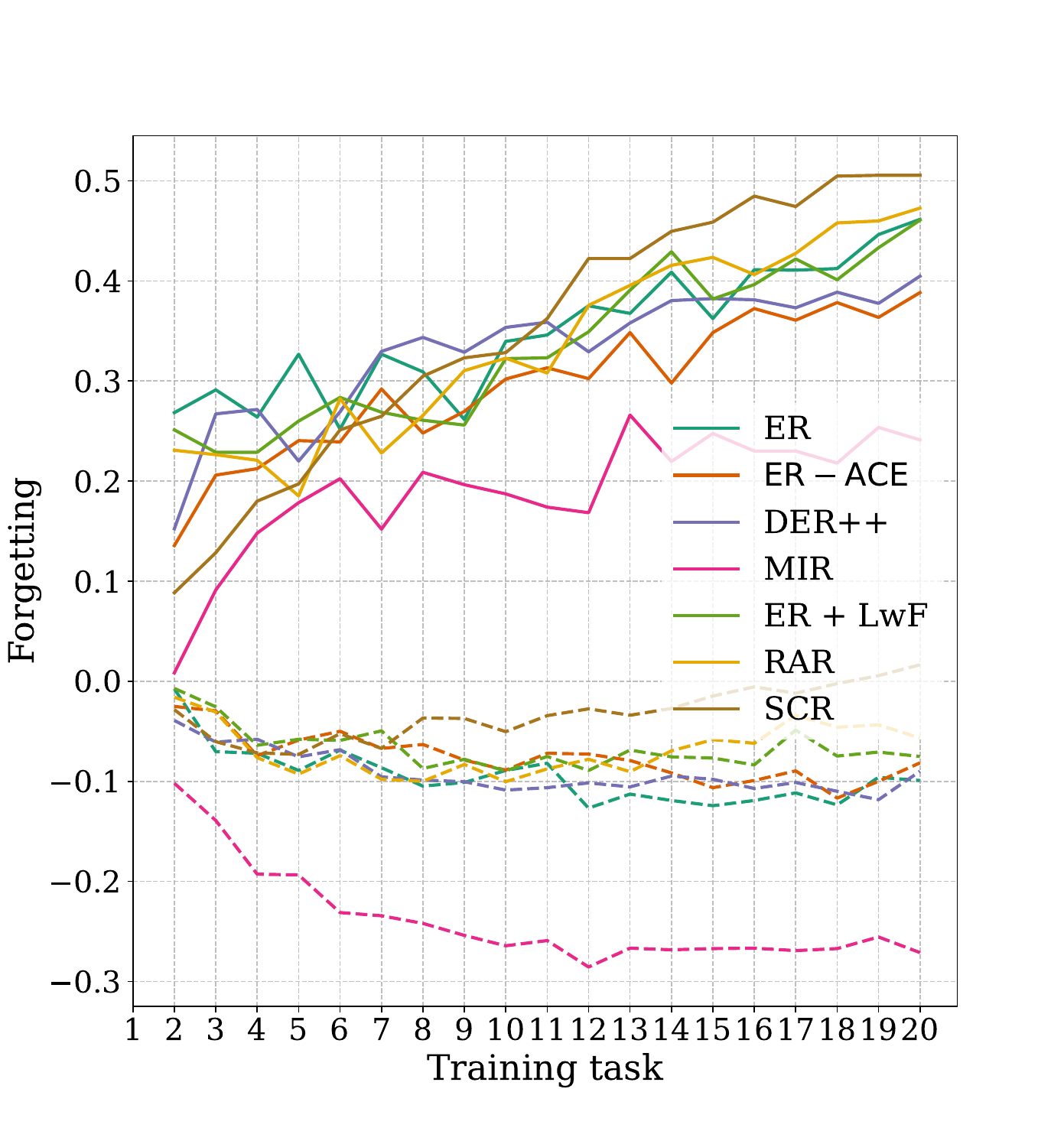}
\caption{#3}
\label{fig:forgetting_cifar100}
\end{figure}
}

\newcommand{\figcifarstab}[3]{
\begin{figure}[tb!]
\centering
\includegraphics[trim=#1, clip, width=#2\linewidth]{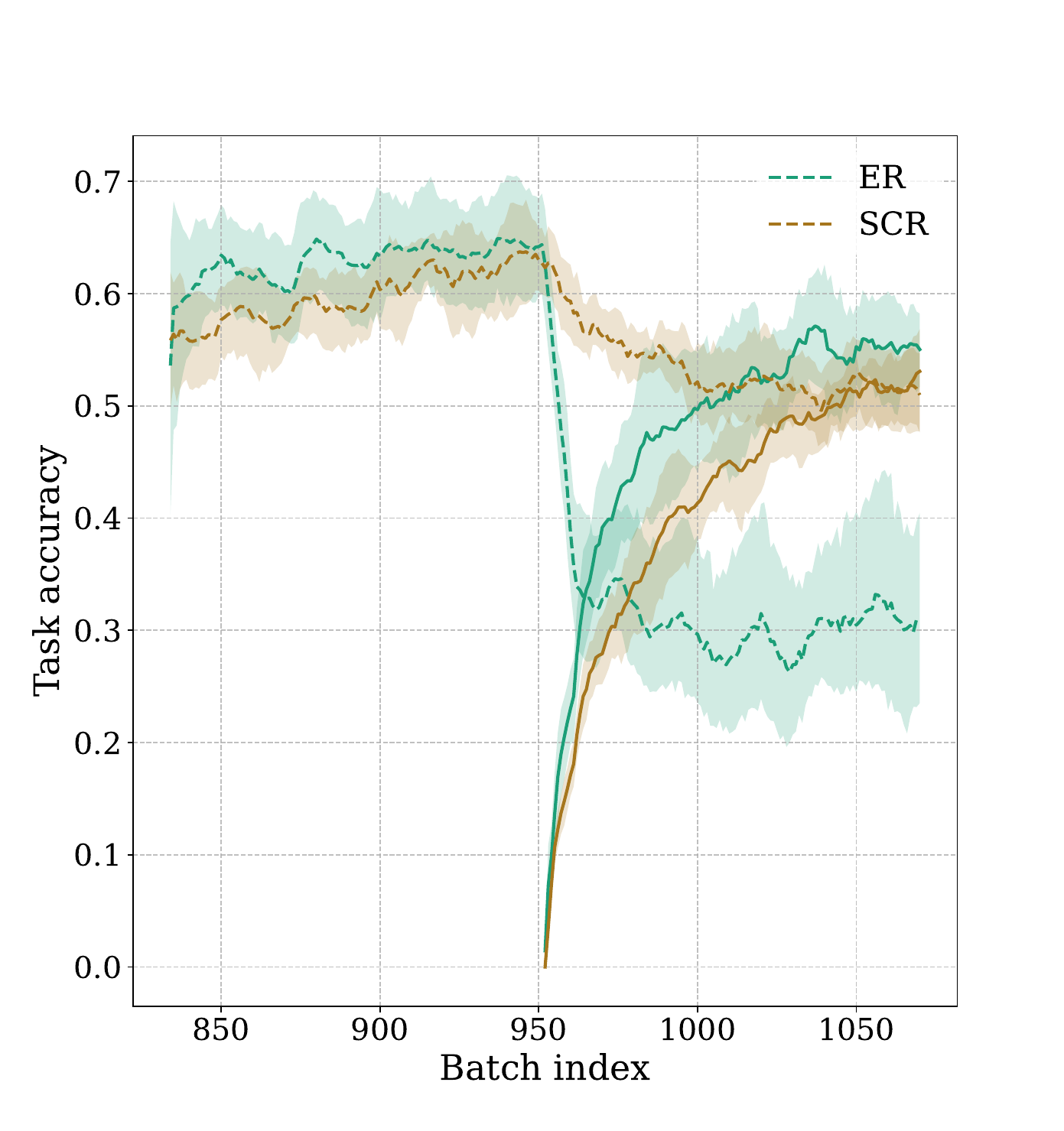}
\caption{#3}
\label{fig:stability_cifar100}
\end{figure}
}

\newcommand{\figcifarmem}[3]{
\begin{figure}[tb!]
\centering
\includegraphics[trim=#1, clip, width=#2\linewidth]{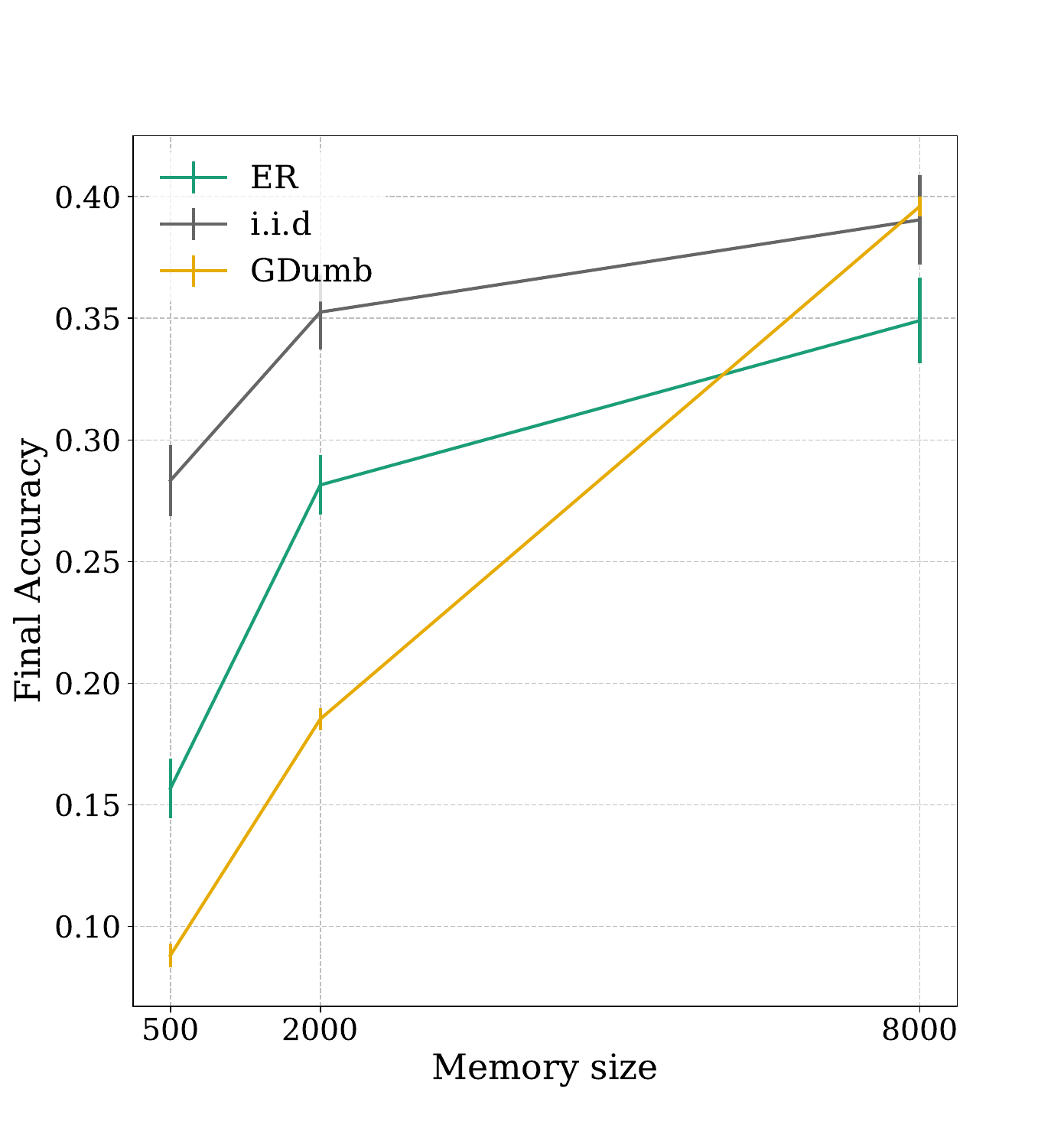}
\caption{#3}
\label{fig:cifar_memory}
\end{figure}
}

\newcommand{\figcifarother}[3]{
\begin{figure*}[htbp]
  \centering
  \begin{subfigure}[b]{0.33\textwidth}
    \centering
    \includegraphics[width=\textwidth]{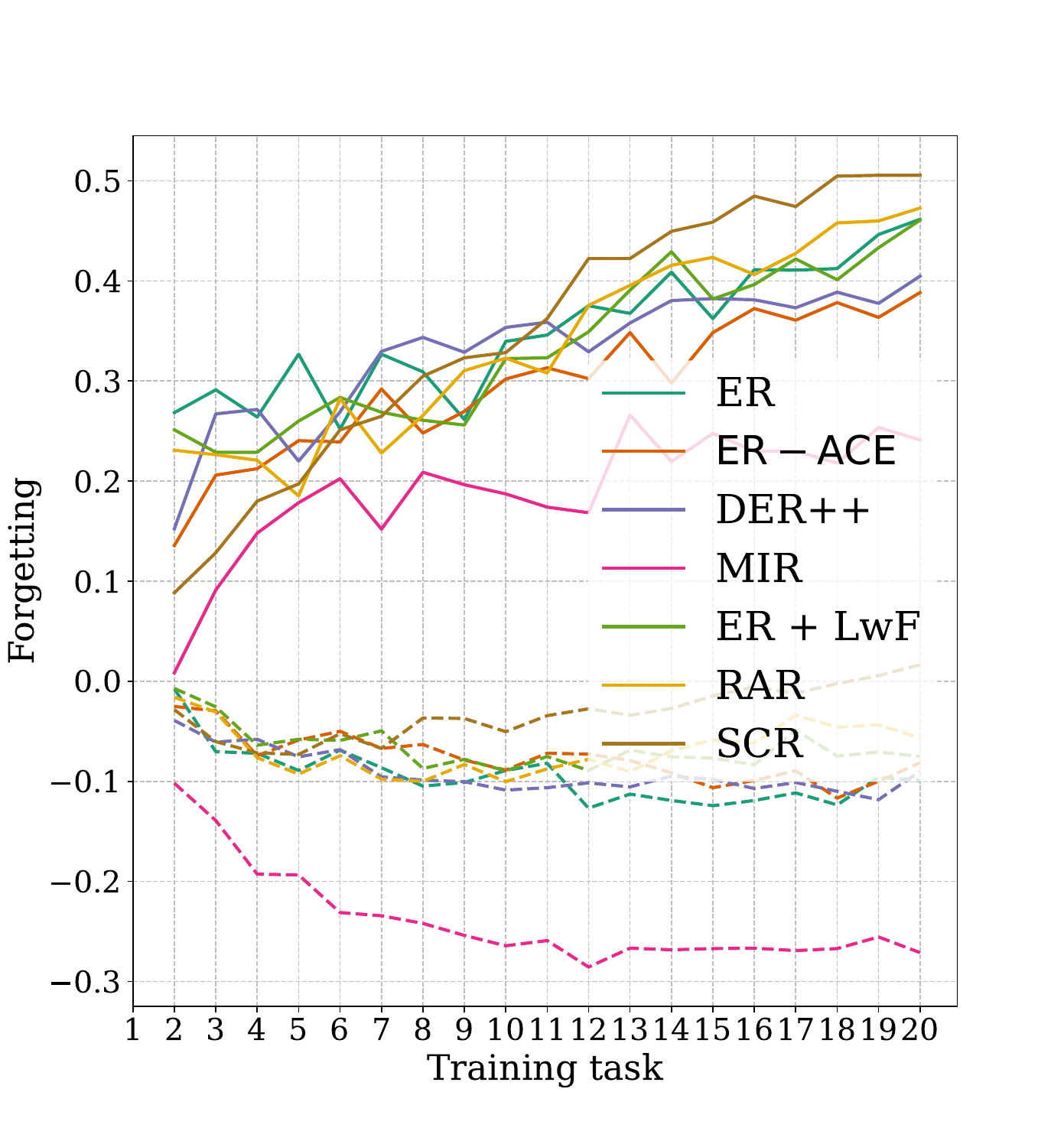}
  \end{subfigure}
  \hfill
  \begin{subfigure}[b]{0.33\textwidth}
    \centering
    \includegraphics[width=\textwidth]{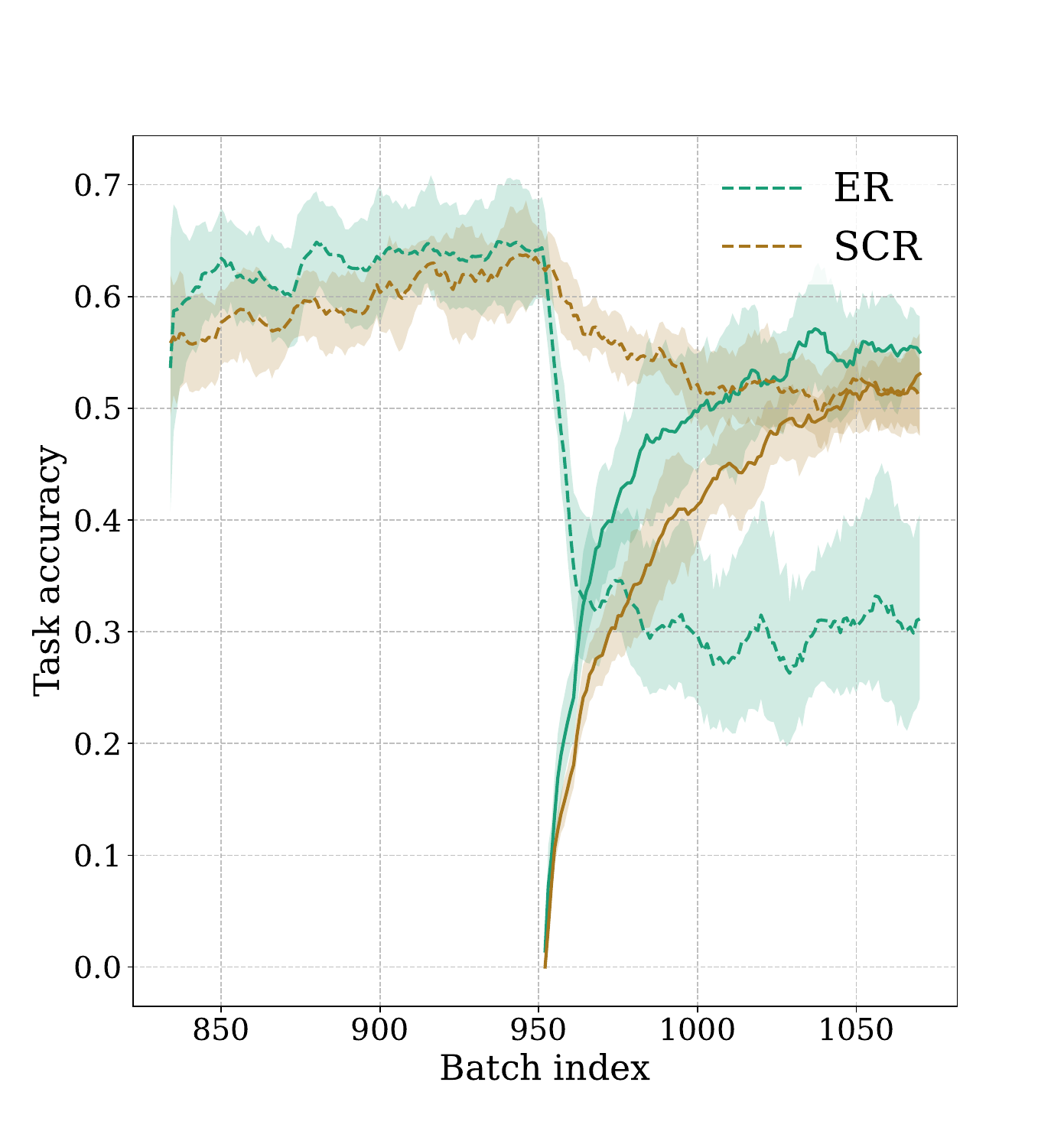}
  \end{subfigure}
  \hfill
  \begin{subfigure}[b]{0.33\textwidth}
    \centering
    \includegraphics[width=\textwidth]{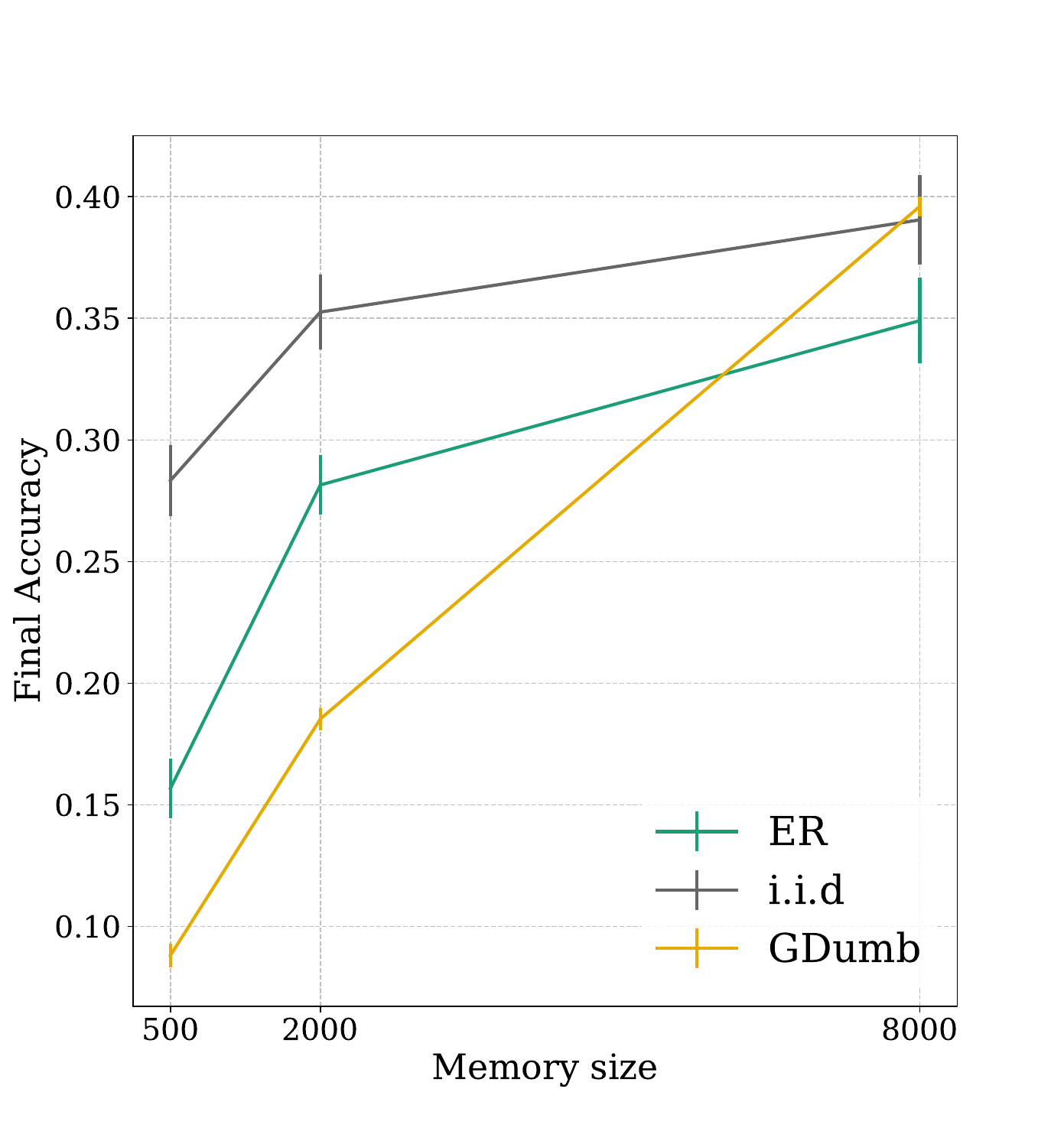}
  \end{subfigure}
  \caption{#3}
  \label{fig:cifar_analysis}
\end{figure*}
}

\newcommand{\tableaccuracycifar}[1]{
\begin{table*}[t]
    \centering
    
    \begin{tabular}{|l|cccc|}
        \hline
        Method & Acc $\uparrow$& $AAA^{val}$ $\uparrow$ & $\operatorname{WC-Acc^{val}}$ $\uparrow$ & Probed Acc $\uparrow$ \\
        \hline
        $i.i.d$ & $35.3 \pm {\scriptstyle 1.5}$ & - & - & 45.8 $\pm$ 0.6 \\
        \hline
        GDumb & 18.5 $\pm$ 0.5 & \text{-} & \text{-} & \text{-} \\
        \hline
        AGEM & 3.1 $\pm$ 0.2 & 10.4 $\pm$ 0.6 & 2.9 $\pm$ 0.3 & 18.7 $\pm$ 0.8 \\
        \hline
        ER & 28.2 $\pm$ 1.2 & 36.6 $\pm$ 2.0 & 12.5 $\pm$ 0.6 &  \textbf{44.9} $\pm$ 0.9 \\
        \hline
        ER + LwF & \textbf{30.4} $\pm$ 0.8 & 39.2 $\pm$ 2.0 & 15.3 $\pm$ 0.9 &  44.4 $\pm$ 0.8 \\
        \hline
        MIR & 29.4 $\pm$ 1.9 & 33.1 $\pm$ 3.2 & 11.6 $\pm$ 1.6 &  43.4 $\pm$ 0.7 \\
        \hline
        ER-ACE & 29.9 $\pm$ 0.6 & 38.5 $\pm$ 1.8 & 14.9 $\pm$ 0.9 &  42.4 $\pm$ 0.6 \\
        \hline
        DER++ & 29.3 $\pm$ 0.9 & 37.5 $\pm$ 2.5 & 13.4 $\pm$ 0.7 &  44.0 $\pm$ 0.8 \\
        \hline
        RAR & 28.2 $\pm$ 1.4 & 38.2 $\pm$ 1.6 & 14.9 $\pm$ 0.7 & 42.3 $\pm$ 0.9 \\
        \hline
        SCR & 28.3 $\pm$ 0.8 & \textbf{42.1} $\pm$ 2.1 & \textbf{20.3} $\pm$ 0.4 &  37.0 $\pm$ 0.3 \\
        \hline
    \end{tabular}
    \vspace{0.3cm}
    \caption{#1}
    \label{tab:accuracies_cifar100}
\end{table*}
}

\newcommand{\tableaccuracytiny}[1]{
\begin{table*}[t]
\centering

\begin{tabular}{|l|cccc|}
\hline
Method & Acc $\uparrow$& $AAA^{val}$ $\uparrow$ & $\operatorname{WC-Acc^{val}}$ $\uparrow$ & Probed Acc $\uparrow$ \\
\hline
$i.i.d$ & 26.5 $\pm$ 0.6 & - & - & 34.3 $\pm$ 0.5 \\
\hline
GDumb & 13.1 $\pm$ 0.4 & \text{-} & \text{-} & \text{-} \\
\hline
AGEM & 2.6 $\pm$ 0.2 & 7.3 $\pm$ 0.5 & 2.6 $\pm$ 0.2 & 23.3 $\pm$ 0.6 \\
\hline
ER & 21.2 $\pm$ 0.6 & 33.9 $\pm$ 1.7 & 15.2 $\pm$ 0.5 &  \textbf{35.6} $\pm$ 0.6 \\
\hline
ER + LwF & 22.7 $\pm$ 1.1 & 34.4 $\pm$ 2.4 & \textbf{17.0} $\pm$ 0.7 &  33.8 $\pm$ 0.9 \\
\hline
MIR & 21.3 $\pm$ 0.8 & 31.0 $\pm$ 1.8 & 15.2 $\pm$ 0.5 & 33.0 $\pm$ 0.4 \\
\hline
ER-ACE & \textbf{23.6} $\pm$ 0.7 & \textbf{35.0} $\pm$ 1.5 & 16.8 $\pm$ 0.7 &  34.2 $\pm$ 0.3 \\
\hline
DER++ & 22.9 $\pm$ 0.5 & 34.2 $\pm$ 4.0 & 16.3 $\pm$ 0.3 & 31.5 $\pm$ 0.9 \\
\hline
RAR & 15.7 $\pm$ 0.9 & 27.8 $\pm$ 2.8 & 10.1 $\pm$ 0.9 &  29.8 $\pm$ 0.9 \\
\hline
SCR & 16.9 $\pm$ 0.4 & 30.7 $\pm$ 1.5 & 12.3 $\pm$ 0.5 &  22.5 $\pm$ 0.4 \\
\hline
\end{tabular}
\vspace{0.3cm}
\caption{#1}
\label{tab:accuracies_tinyimagenet}
\end{table*}
}

\newcommand{\tabletemplate}[1]{
\begin{table*}[tb!]
\centering

\begin{tabular}{|l|cccc|}
\hline
Method & Acc $\uparrow$& $AAA^{val}$ $\uparrow$ & $\operatorname{WC-Acc^{val}}$ $\uparrow$ & Probed Acc $\uparrow$ \\
\hline
$i.i.d$ & TODO $\pm$ TODO & - & - & TODO $\pm$ TODO \\
\hline
GDumb & TODO $\pm$ TODO & \text{-} & \text{-} & \text{-} \\
\hline
AGEM & TODO $\pm$ TODO & TODO $\pm$ TODO & TODO $\pm$ TODO & TODO $\pm$ TOOD \\
\hline
ER & TODO $\pm$ TODO & TODO $\pm$ TODO & TODO $\pm$ TODO &  TODO $\pm$ TODO \\
\hline
ER + LwF & TODO $\pm$ TODO & TODO $\pm$ TODO & TODO $\pm$ TODO &  TODO $\pm$ TODO \\
\hline
MIR & TODO $\pm$ TODO & TODO $\pm$ TODO & TODO $\pm$ TODO &  TODO $\pm$ TODO \\
\hline
ER-ACE & TODO $\pm$ TODO & TODO $\pm$ TODO & TODO $\pm$ TODO &  TODO $\pm$ TODO \\
\hline
DER++ & TODO $\pm$ TODO & TODO $\pm$ TODO & TODO $\pm$ TODO &  TODO $\pm$ TODO \\
\hline
RAR & TODO $\pm$ TODO & TODO $\pm$ TODO & TODO $\pm$ TODO &  TODO $\pm$ TODO \\
\hline
SCR & TODO $\pm$ TODO & TODO $\pm$ TODO & TODO $\pm$ TODO &  TODO $\pm$ TODO \\
\hline
\end{tabular}
\vspace{0.3cm}
\caption{#1}
\label{tab:template}
\end{table*}
}

\newcommand*{\recommendbox}[1]{\colorbox{light-gray}{\parbox{.98\linewidth}{#1}}}
\newcommand{\tablerecommendations}{
\begin{table*}
\label{tab:recommendations}
\recommendbox{
\centering
\vspace{0.1cm}
\begin{tabular}{p{0.45\linewidth}p{0.49\linewidth}}
\multicolumn{2}{c}{\textsc{Major findings of our performance evaluation on Online Continual Learning}\vspace{0.1cm}}\\
\begin{minipage}[t]{\linewidth} 
  \begin{itemize}
    \item Good stability does not necessarily transfer to higher accuracy (See Table \ref{tab:accuracies}, Figure \ref{fig:cifar_analysis} and Section \ref{sec:results} \textbf{Stability}).
    \item There is no best-performing OCL method across all metrics or memory sizes (See Table \ref{tab:accuracies}).
    \item OCL methods suffer from under-fitting in the common experimental setup (See Figure \ref{fig:cifar_analysis} and Section \ref{sec:results} \textbf{Forgetting}).
  \end{itemize} 
\end{minipage} &
\begin{minipage}[t]{\linewidth} 
  \begin{itemize}
    \item Well properly tuned ER is a very competitive baseline obtaining better results than most existing methods (See \textbf{memory batch size} discussion \ref{sec:results} and Section \ref{sec:setup} \textbf{Implementation}).
    \item The quality of the representation is very close to the one learned on the i.i.d stream, indicating that learning a good classifier is one of main problems. (See Section \ref{sec:results} \textbf{Representation quality}).  
  \end{itemize} 
\end{minipage} \\
\end{tabular}
\vspace{0.1cm}
}
\end{table*}
}

\newcommand{\tablecifarsmall}[1]{
\renewcommand{\arraystretch}{1.1} 
\begin{table*}[tb!]
\centering

\begin{tabular}{l@{} c cccc}
\Xhline{1.3pt}
\textbf{Method} & & Acc $\uparrow$& $AAA^{val}$ $\uparrow$ & $\operatorname{WC-Acc^{val}}$ $\uparrow$ & Probed Acc $\uparrow$ \\
\cline{1-1} \cline{3-6} 
$i.i.d$ & & $28.3 \pm {\scriptstyle 1.5}$ & - & - & $40.0 \pm {\scriptstyle  0.9}$ \\
\cline{1-1} \cline{3-6} 
GDumb & & $8.8 \pm {\scriptstyle 0.5}$ & \text{-} & \text{-} & \text{-} \\
\cline{1-1} \cline{3-6} 
AGEM & & $3.2 \pm {\scriptstyle  0.4}$ & $10.4 \pm {\scriptstyle  0.5}$ & $3.2 \pm {\scriptstyle 0.3}$ & $19.2 \pm {\scriptstyle 0.7}$ \\
\cline{1-1} \cline{3-6} 
ER & & $15.7 \pm {\scriptstyle 1.2}$ & $28.6 \pm {\scriptstyle  1.7}$ & $7.7 \pm {\scriptstyle  0.9}$ & $\mathbf{38.2} \pm {\scriptstyle  1.2}$ \\
\cline{1-1} \cline{3-6} 
ER + LwF & & $19.7 \pm {\scriptstyle  1.5} $ & $32.5 \pm {\scriptstyle  1.9}$ & $10.6 \pm {\scriptstyle  0.9}$ &  $ 38.0 \pm {\scriptstyle  1.6}$ \\
\cline{1-1} \cline{3-6} 
MIR & & $15.7 \pm {\scriptstyle 1.4}$ & $27.4 \pm {\scriptstyle  2.4}$ & $9.3 \pm {\scriptstyle  7.7}$ &  $36.2 \pm {\scriptstyle  1.0}$ \\
\cline{1-1} \cline{3-6} 
ER-ACE & & $\mathbf{20.8} \pm {\scriptstyle  0.9}$ & $\mathbf{32.8} \pm {\scriptstyle  2.2}$ & $\mathbf{11.5} \pm {\scriptstyle  0.5}$ &  $36.8 \pm {\scriptstyle  1.1}$ \\
\cline{1-1} \cline{3-6} 
DER++ & & $ 15.2 \pm {\scriptstyle  1.4}$  & $ 28.9 \pm {\scriptstyle  3.0}$  & $ 7.9 \pm {\scriptstyle  0.6}$ &  $ 37.1 \pm {\scriptstyle  1.5}$ \\
\cline{1-1} \cline{3-6} 
RAR & & $14.6 \pm {\scriptstyle  1.2}$ & $28.6 \pm {\scriptstyle 1.5}$ & $7.9 \pm {\scriptstyle 0.6}$ & $35.7 \pm {\scriptstyle 0.9}$ \\
\cline{1-1} \cline{3-6} 
SCR & & $13.2 \pm {\scriptstyle 0.5}$ & $29.4 \pm {\scriptstyle 1.9}$ & $8.5 \pm {\scriptstyle 0.5}$ & $28.4 \pm {\scriptstyle 0.5}$ \\
\Xhline{1.3pt}
\end{tabular}

\vspace{0.3cm}
\caption{#1}
\label{tab:cifar500}
\end{table*}
}

\newcommand{\tablecifarbig}[1]{
\renewcommand{\arraystretch}{1.1} 
\begin{table*}[tb!]
\centering

\begin{tabular}{l@{} c cccc}
    \Xhline{1.3pt}
    \textbf{Method} & & Acc $\uparrow$& $AAA^{val}$ $\uparrow$ & $\operatorname{WC-Acc^{val}}$ $\uparrow$ & Probed Acc $\uparrow$ \\
    \cline{1-1} \cline{3-6} 
    $i.i.d$ & & $39.0 \pm {\scriptstyle 1.8}$ & - & - & $49.3 \pm {\scriptstyle 0.9}$ \\
    \cline{1-1} \cline{3-6} 
    GDumb & & $39.6 \pm {\scriptstyle 0.4}$ & \text{-} & \text{-} & \text{-} \\
    \cline{1-1} \cline{3-6} 
    AGEM & & $3.1 \pm {\scriptstyle 0.3}$ & $10.5 \pm {\scriptstyle 0.5}$ & $3.1 \pm {\scriptstyle 0.2}$ & $18.6 \pm {\scriptstyle 0.8}$ \\
    \cline{1-1} \cline{3-6} 
    ER & & $34.9 \pm {\scriptstyle 1.8}$ & $39.1 \pm {\scriptstyle 1.7}$ & $13.2 \pm {\scriptstyle 0.8}$ &  $48.7 \pm {\scriptstyle 0.7}$ \\
    \cline{1-1} \cline{3-6} 
    ER + LwF & & $36.7 \pm {\scriptstyle 1.3}$ & $41.7 \pm {\scriptstyle 1.8}$ & $17.2 \pm {\scriptstyle 0.9}$ & $ 48.5 \pm {\scriptstyle 0.9}$ \\
    \cline{1-1} \cline{3-6} 
    MIR & & $31.8 \pm {\scriptstyle 1.4}$ & $33.6 \pm {\scriptstyle 2.6}$ & $8.4 \pm {\scriptstyle 1.4}$ & $47.8 \pm {\scriptstyle 0.8}$ \\
    \cline{1-1} \cline{3-6} 
    ER-ACE & & $35.1 \pm {\scriptstyle 1.2}$ & $40.6 \pm {\scriptstyle 1.5}$ & $16.8 \pm {\scriptstyle 1.1}$ & $47.0 \pm {\scriptstyle 0.7}$ \\
    \cline{1-1} \cline{3-6} 
    DER++ & & $ 36.1 \pm {\scriptstyle 1.7}$ & $40.8 \pm {\scriptstyle 2.0}$ & $ 14.6 \pm {\scriptstyle 0.4}$ &  $49.1 \pm {\scriptstyle 0.7}$ \\
    \cline{1-1} \cline{3-6} 
    RAR & & $36.9 \pm {\scriptstyle 2.0}$ & $42.2 \pm {\scriptstyle 1.3}$ & $16.1 \pm {\scriptstyle 1.2}$ & $48.1 \pm {\scriptstyle 1.2}$ \\
    \cline{1-1} \cline{3-6} 
    SCR & & $\mathbf{43.5} \pm {\scriptstyle 0.7}$ & $50.2 \pm {\scriptstyle 2.1}$ & $\mathbf{32.6} \pm {\scriptstyle 0.7}$ & $47.3 \pm {\scriptstyle 0.7}$ \\
    \cline{1-1} \cline{3-6} 
    ER\footnotemark & & $43.0 \pm {\scriptstyle 0.7}$ & $\mathbf{52.7}  \pm {\scriptstyle 2.2}$ & $30.7 \pm {\scriptstyle 0.9}$ & $\mathbf{49.5} \pm {\scriptstyle 1.4}$ \\
    \Xhline{1.3pt}
\end{tabular}
\vspace{0.3cm}
\caption{#1}
\label{tab:cifar8k}
\end{table*}
\footnotetext[2]{Modified ER version with a memory batch size of size 118 (to match the size of the SCR one)}
}

\newcommand{\tableruntime}[1]{
\begin{table}[tb!]
\centering

\resizebox{0.8\textwidth}{!}{%
\begin{tabular}{|l|c|}
\hline
Method & \\

\end{tabular}
}
\vspace{0.3cm}
\caption{#1}
\label{tab:runtime}
\end{table*}
}

\newcommand{\tablemerged}[1]{
\renewcommand{\arraystretch}{1.1} 
\begin{table*}[t]
    \centering
    \resizebox{1.0\textwidth}{!}{%
    \begin{tabular}{l@{} c cccc c@{} cccc}
        \Xhline{1.3pt}
        \textbf{Method} & & \multicolumn{4}{c }{\textbf{Split-Cifar100 (20 Tasks)}} & & \multicolumn{4}{c}{\textbf{Split-TinyImagenet (20 Tasks)}} \\
         & & Acc $\uparrow$& $AAA^{val}$ $\uparrow$ & $\operatorname{WC-Acc^{val}}$ $\uparrow$ & Probed Acc $\uparrow$ &  & Acc $\uparrow$& $AAA^{val}$ $\uparrow$ & $\operatorname{WC-Acc^{val}}$ $\uparrow$ & Probed Acc $\uparrow$ \\
        
        \cline{1-1} \cline{3-6} \cline{8-11}
        $i.i.d$ & & $35.3 \pm {\scriptstyle 1.5}$ & - & - & $45.8 \pm {\scriptstyle 0.6}$ & & $26.5 \pm {\scriptstyle 0.6}$ & - & - & $34.3 \pm {\scriptstyle 0.5}$ \\
        \cline{1-1} \cline{3-6} \cline{8-11}
        GDumb & & $18.5 \pm {\scriptstyle 0.5}$ & \text{-} & \text{-} & \text{-} & & $13.1 \pm {\scriptstyle 0.4}$ & \text{-} & \text{-} & \text{-} \\
        \cline{1-1} \cline{3-6} \cline{8-11}
        AGEM & & $3.1 \pm {\scriptstyle 0.2}$ & $ 10.4 \pm {\scriptstyle 0.6} $ & $ 2.9 \pm {\scriptstyle 0.3} $ & $18.7 \pm {\scriptstyle 0.8} $ & & $ 2.6 \pm {\scriptstyle 0.2} $ & $7.3 \pm {\scriptstyle 0.5}$ & $2.6 \pm {\scriptstyle 0.2} $ & $23.3 \pm {\scriptstyle 0.6}$ \\
        \cline{1-1} \cline{3-6} \cline{8-11}
        ER & & $28.2 \pm {\scriptstyle 1.2}$ & $36.6 \pm {\scriptstyle 2.0}$ & $ 12.5 \pm {\scriptstyle 0.6}$ &  $\mathbf{44.9} \pm {\scriptstyle 0.9}$ & & $21.2 \pm {\scriptstyle 0.6}$ & $33.9 \pm {\scriptstyle 1.7}$ & $15.2 \pm {\scriptstyle 0.5}$ &  $\mathbf{35.6} \pm {\scriptstyle 0.6}$ \\
        \cline{1-1} \cline{3-6} \cline{8-11}
        ER + LwF & & $\mathbf{30.4} \pm {\scriptstyle 0.8}$ & $39.2 \pm {\scriptstyle 2.0}$ & $15.3 \pm {\scriptstyle 0.9}$ &  $44.4 \pm {\scriptstyle 0.8}$ & & $22.7 \pm {\scriptstyle 1.1}$ & $34.4 \pm {\scriptstyle 2.4}$ & $\textbf{17.0} \pm {\scriptstyle 0.7}$ &  $33.8 \pm {\scriptstyle 0.9}$ \\
        \cline{1-1} \cline{3-6} \cline{8-11}
        MIR & & $29.4 \pm {\scriptstyle 1.9}$ & $33.1 \pm {\scriptstyle 3.2}$ & $ 11.6 \pm {\scriptstyle 1.6} $ &  $43.4 \pm {\scriptstyle 0.7}$ & & $ 21.3 \pm {\scriptstyle 0.8}$ & $ 31.0 \pm {\scriptstyle 1.8}$ & $15.2 \pm {\scriptstyle 0.5}$ & $33.0 \pm {\scriptstyle 0.4} $ \\
        \cline{1-1} \cline{3-6} \cline{8-11}
        ER-ACE & & $29.9 \pm {\scriptstyle 0.6}$ & $38.5 \pm {\scriptstyle 1.8}$ & $14.9 \pm {\scriptstyle 0.9}$ &  $42.4 \pm {\scriptstyle 0.6}$ & & $\mathbf{23.6} \pm {\scriptstyle 0.7}$ & $\mathbf{35.0} \pm {\scriptstyle 1.5} $ & $ 16.8 \pm {\scriptstyle 0.7} $ &  $ 34.2 \pm  {\scriptstyle 0.3}$ \\
        \cline{1-1} \cline{3-6} \cline{8-11}
        DER++ & & $29.3 \pm {\scriptstyle 0.9}$ & $37.5 \pm {\scriptstyle 2.5}$  & $ 13.4 \pm {\scriptstyle 0.7} $ &  $44.0 \pm {\scriptstyle 0.8} $  & & $ 22.9 \pm {\scriptstyle 0.5} $  & $34.2 \pm {\scriptstyle 4.0} $  & $ 16.3 \pm {\scriptstyle 0.3} $  & $ 31.5 \pm  {\scriptstyle 0.9} $ \\
        \cline{1-1} \cline{3-6} \cline{8-11}
        RAR & & $28.2 \pm {\scriptstyle 1.4}$ & $38.2 \pm {\scriptstyle 1.6} $ & $ 14.9 \pm {\scriptstyle 0.7} $  & $ 42.3 \pm {\scriptstyle 0.9} $  & & $ 15.7 \pm {\scriptstyle 0.9}  $  & $ 27.8 \pm {\scriptstyle 2.8} $  & $10.1 \pm {\scriptstyle 0.9} $ &  $ 29.8 \pm {\scriptstyle 0.9}$  \\
        \cline{1-1} \cline{3-6} \cline{8-11}
        SCR & & $28.3 \pm {\scriptstyle 0.8} $ & $\mathbf{42.1} \pm {\scriptstyle 2.1}$ & $\mathbf{20.3} \pm {\scriptstyle 0.4}$ &  $ 37.0 \pm {\scriptstyle 0.3}$  & & $ 16.9 \pm {\scriptstyle 0.4}$ & $ 30.7 \pm {\scriptstyle 1.5}$  & $ 12.3 \pm {\scriptstyle 0.5}$  &  $ 22.5 \pm {\scriptstyle 0.4}$ \\
        \Xhline{1.3pt}
    \end{tabular}
    }
    \vspace{0.3cm}
    \caption{#1}
    \label{tab:accuracies}
\end{table*}
}

\title{A Comprehensive Empirical Evaluation on Online Continual Learning}


\author{Albin Soutif--Cormerais\\
Computer Vision Center\\
Universitat Autònoma de Barcelona\\
Barcelona, Spain\\
{\tt\small albin@cvc.uab.cat}
\and
Antonio Carta\\
Department of Computer Science\\
University of Pisa\\
Pisa, Italy\\
{\tt\small antonio.carta@unipi.it}
\and
Andrea Cossu\\
$\qquad$ Scuola Normale Superiore $\qquad$\\
Pisa, Italy\\
{\tt\small andrea.cossu@sns.it}
\and
Julio Hurtado\\
Department of Computer Science\\
University of Pisa\\
Pisa, Italy\\
{\tt\small julio.hurtado@di.unipi.it}
\and
Hamed Hemati\\
$\qquad$ University of St. Gallen  $\qquad$\\
Saint-Gallen, Switzerland\\
{\tt\small hamed.hemati@unisg.ch}
\and
Vincenzo Lomonaco\\
Department of Computer Science\\
University of Pisa\\
Pisa, Italy\\
{\tt\small vincenzo.lomonaco@unipi.it}
\and
Joost van de Weijer\\
Computer Vision Center\\
Universitat Autònoma de Barcelona\\
Barcelona, Spain\\
{\tt\small joost@cvc.uab.cat}
}

\maketitle
\ificcvfinal\thispagestyle{empty}\fi

\begin{abstract}

Online continual learning aims to get closer to a live learning experience by learning directly on a stream of data with temporally shifting distribution and by storing a minimum amount of data from that stream. In this empirical evaluation, we evaluate various methods from the literature that tackle online continual learning. More specifically, we focus on the class-incremental setting in the context of image classification, where the learner must learn new classes incrementally from a stream of data. We compare these methods on the Split-CIFAR100 and Split-TinyImagenet benchmarks, and measure their average accuracy, forgetting, stability, and quality of the representations, to evaluate various aspects of the algorithm at the end but also during the whole training period. We find that most methods suffer from stability and underfitting issues. However, the learned representations are comparable to i.i.d. training under the same computational budget. No clear winner emerges from the results and basic experience replay, when properly tuned and implemented, is a very strong baseline. We release our modular and extensible codebase at \url{https://github.com/AlbinSou/ocl_survey} based on the avalanche framework to reproduce our results and encourage future research.


\end{abstract}

\section{Introduction}



In recent years, we have witnessed a surge of interest and progress in deep continual learning methodologies. These methods are able to learn continually from a stream of non-stationary data, thereby relaxing the principal assumption of having access to \emph{independent and identically distributed} (\emph{i.i.d.}) samples, often made in statistical learning \cite{hastie2009elements}. In classic (or batch) continual learning, the common assumption is that the data stream is composed of distinct, explicitly defined \emph{tasks} or \emph{domains}, and that the method can detect the task boundaries or easily switch between domains. However, in many real-world scenarios, the data stream may not have clear task boundaries or domain labels, and the method may need to quickly adapt to changes in the data distribution \cite{aljundi2019gradient, koh2022online}. Moreover, these methods have access to a batch of data at each task, normally in the form of a dataset, providing them with a locally i.i.d. access to the data.


\emph{Online Continual Learning} (OCL)~\cite{lopez2017gradient} is a more challenging and realistic setting of continual learning, where similar to online learning \cite{mohri2018foundations}, the method learns from \emph{each arriving data point} in the stream. OCL methods \emph{update the model with a high frequency}, often without access to task labels or boundaries, and with a limited computational and memory budget. Due to the non-stationary nature of the stream, OCL methods need to balance stability and plasticity. Furthermore, since the model is used for inference at each time step (\emph{anytime inference}), the learning algorithm must not suffer from stability issues at any point of the learning process.  

State-of-the-art OCL methods use a rehearsal buffer to mitigate forgetting~\cite{chaudhry2019tiny}. Several improvements of basic experience replay have been proposed to improve the sampling from the buffer~\cite{MIR,kumari2022retrospective}, the loss function~\cite{caccia2021new,buzzega2020dark,mai2021supervised}, weights update~\cite{chaudhry2018efficient} or the classification layer~\cite{mai2021supervised}. A common limitation of these works, and even recent empirical surveys~\cite{mai2022online} is that they focus only on forgetting and final accuracy. However, OCL methods have several other objectives that should be measured with appropriate metrics.

To encourage progress in Online Continual Learning, in this paper, we provide a \emph{comprehensive empirical evaluation} of OCL methods. We exploit recent proposals that provide better metrics to measure forgetting~\cite{soutif2021importance}, continual stability~\cite{lange2023continual}, and quality of the representations \cite{davari2022probing, caccia2021new, koh2022online}. The experimental results on these metrics highlight the strength and limitations of state-of-the-art methods in OCL. 

In this work, we contribute to the body of literature in this area as follows:

\tablerecommendations

\begin{itemize}
    \item We formally define \emph{Online Continual Learning} (Section \ref{sec:ocl_scenario}) and a comprehensive set of metrics (Section \ref{sec:metrics}) to measure the performance across different dimensions: accuracy, forgetting, continual stability, and quality of the latent representations. 
    \item We conduct a comprehensive empirical evaluation on \emph{Class-Incremental} scenarios on two main benchmarks (\emph{Split-CIFAR100} and \emph{Split-TinyImagenet}) evaluating 9 different approaches against 5 metrics. The main findings can be found in the “recommendations box” at the top of this page.
    \item We release all the code to reproduce our results, compare and easily prototype new OCL strategies. The code has been developed within the Deep Continual Learning Avalanche library for maximum flexibility and reproducibility.
\end{itemize}

\section{Online Continual Learning}\label{sec:ocl_scenario}

In \emph{Online Continual Learning (OCL)}, a model learns from a stream of non-stationary experiences of data. In a classification problem, at each timestep $t$ a mini-batch $(x_t, y_t) \sim p_t(x, y)$ is available, where $p_t$ is the underlying distribution of data and may change over time. The function $f: \mathbb{R}^n \rightarrow \mathbb{R}^c$ is the prediction function and maps inputs to the unnormalized class probabilities (logits). An OCL algorithm is a function $\mathcal{A}: (x_t, y_t), f_{t-1}, \mathcal{M}_{t-1} \rightarrow f_{t}, \mathcal{M}_{t}$, where $f_t$ is the model at time $t$ and $\mathcal{M}_t = \{ (x_i, y_i) \} $ is a replay buffer, {\it i.e.}, a small set of samples from the past stream stored for rehearsal.

Unlike Offline Continual Learning, OCL is a challenging setting where only a few new samples are available at each step, which can be stored in a very limited amount. The following are some properties that characterize OCL setups:
\paragraph{Online.} At each timestep only a small mini-batch $(x_t, y_t)$ is available (10 in our experiments).
\paragraph{Task Labels.} In a task-aware setting, the models know that samples belong to a set of known tasks and a task label is available to associate each sample to its own task. We assume a task-agnostic setting where task labels are not available.
\paragraph{Task Boundaries.} Even in the absence of task labels, many continual learning methods assume knowledge about task boundaries, expecting to know when the data distribution switches to a new task. In a boundary-agnostic setting, this information is not available\footnote{This is commonly referred to as task-free, and it is a common assumption in OCL. We propose the term boundary-agnostic to highlight that task labels and boundaries are two different kinds of knowledge about the properties of the stream}. In this paper, we test both boundary-agnostic and boundary-aware methods.
\paragraph{Anytime Inference.} In most OCL applications, models should be able to train but also to perform inference online, after every training step \cite{koh2022online}. As a result, methods that require expensive finetuning steps before inference such as GDumb \cite{prabhu2020gdumb} are not considered OCL methods in this paper.


\section{Methods}

\label{sec:methods}


\renewcommand{\arraystretch}{1.1} 
\begin{table}[h!]
\resizebox{0.48\textwidth}{!}{
    \centering
    \begin{tabular}{c@{} c c@{} c c@{} c c@{} c}
         \Xhline{1.3pt}
         \textbf{Name} & & \textbf{Elements} &  & \textbf{Year} & & \textbf{Boundary} \\
         \cline{1-1} \cline{3-3}  \cline{5-5} \cline{7-7}  
         AGEM \cite{chaudhry2018efficient} & & Modified Update & & 2018 & &  \\
         \cline{1-1} \cline{3-3}  \cline{5-5} \cline{7-7}  
         ER \cite{chaudhry2019tiny} & & - & & 2019 & &  \\
         \cline{1-1} \cline{3-3}  \cline{5-5} \cline{7-7}  
         ER + LwF \cite{li2017learning} & & Distillation Loss & & 2019 & & \checkmark \\
         \cline{1-1} \cline{3-3}  \cline{5-5} \cline{7-7}  
         ER-ACE \cite{caccia2021new} & & Modified Cross-Entropy Loss & & 2021 & & \\
         \cline{1-1} \cline{3-3}  \cline{5-5} \cline{7-7}  
         MIR \cite{MIR}  & & Modified sampling & & 2019 & & \\
         \cline{1-1} \cline{3-3}  \cline{5-5} \cline{7-7}  
         SCR \cite{mai2021supervised} & & Contrastive Loss, NMC & & 2021 & &  \\
         \cline{1-1} \cline{3-3}  \cline{5-5} \cline{7-7}  
         RAR \cite{kumari2022retrospective} & & Adversarial Augmentations & & 2022 & & \\
         \cline{1-1} \cline{3-3}  \cline{5-5} \cline{7-7}  
         DER++ \cite{buzzega2020dark} & & Distillation Loss & & 2020 & & \\
         \cline{1-1} \cline{3-3}  \cline{5-5} \cline{7-7}   
         GDumb \cite{prabhu2020gdumb} & & Offline finetuning on the buffer &  & 2020 &  & \checkmark \\
         \Xhline{1.3pt}
    \end{tabular}
}
    \vspace{0.3cm}
    \caption{Summary of methods tried in the survey along with their particularities (release year, access to task boundaries).}
    \label{tab:method}
\end{table}

\begin{figure*}
    \centering
    \includegraphics[width=\linewidth]{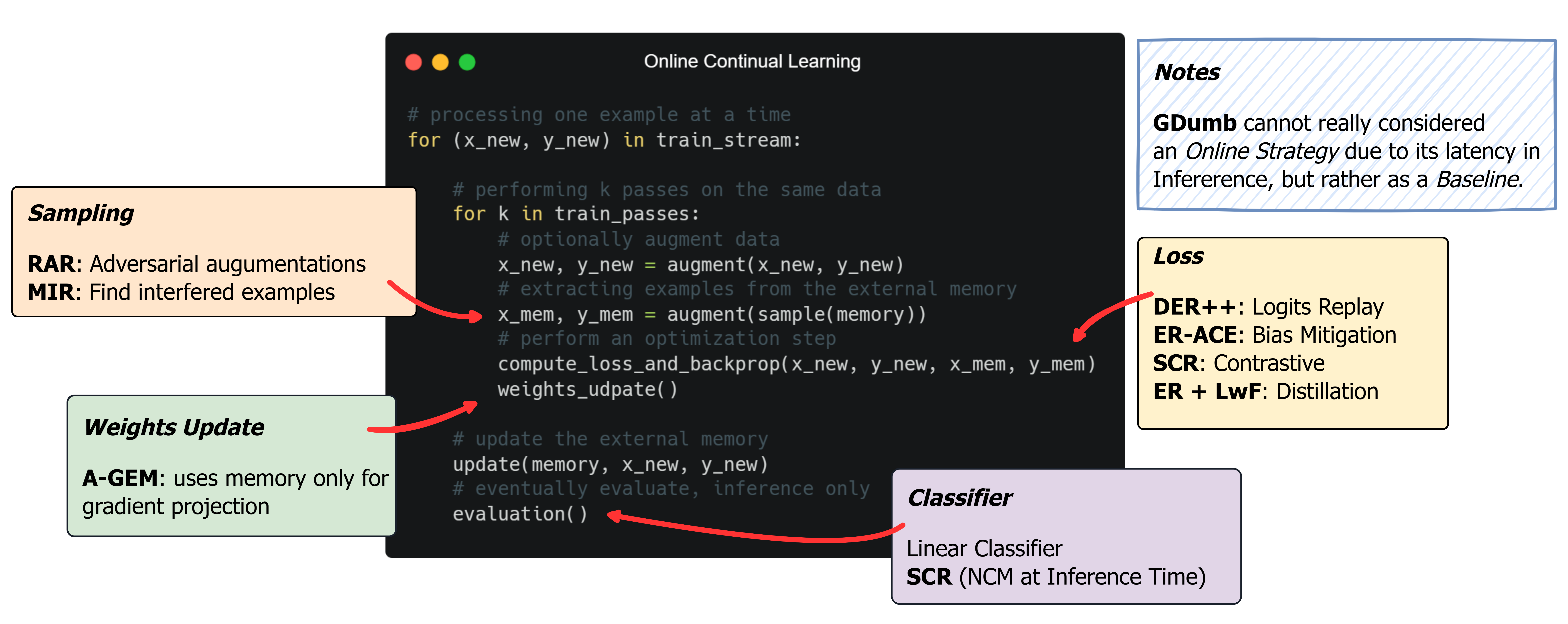}
    \caption{Pseudocode of replay-based OCL methods. Each method can be obtained from basic experience replay (ER) by modifying one of its fundamental components: sampling, loss, classifier, weight update.}
    \label{fig:ocl-algo}
\end{figure*}

\paragraph{} Similar to Offline Continual Learning, in OCL, most state-of-the-art results are obtained by rehearsal-based methods. Because of this, most approaches use rehearsal, which is the main reason we focus on them for our empirical study. Rehearsal-based methods keep a separate memory $\mathcal{M}$ of fixed size to store past samples, updated after each mini-batch. Most rehearsal-based approaches, and all the chosen methods, follow the pseudocode shown in Figure~\ref{fig:ocl-algo}. In the following paragraphs, we will describe in detail each line, explaining some methods-specific additions and the main reason for selecting each approach.

\paragraph{Sampling.} Usually, each new mini-batch is reused for multiple training passes, each time sampling a different mini-batch from the memory and applying stochastic augmentations to both old and new data. This is justified by the theoretical analysis in \cite{zhang2022simple}. MIR \cite{MIR} finds the maximally interfered samples, {\it i.e.}, those that maximally decrease their loss after an SGD step on the new data, and select those for rehearsal. Instead, RAR~\cite{kumari2022retrospective} generates new samples using targeted adversarial attacks that are designed to be in close proximity to the decision boundary of the classifier.

\paragraph{Loss.} Most methods use a supervised loss on both new and memory data, such as the cross-entropy. ER-ACE \cite{caccia2021new} use different loss on new and old data due to the different nature of the two samples. DER++ \cite{buzzega2020dark} uses logits distillation, storing the logits together with the raw data in the memory. While the objective is a knowledge distillation loss, the teacher used to compute the logits depend on the time when the sample was stored. SCR \cite{mai2021supervised} uses a contrastive loss. In CL settings with large batches, it was shown that contrastive losses suffer less from forgetting than supervised ones~\cite{cossuContinualPreTrainingMitigates2022,madaanRepresentationalContinuityUnsupervised2022,finiSelfSupervisedModelsAre2022}. However, many contrastive losses require large batch sizes, big and diverse datasets, and more time to converge, which may not be possible in an online setting. Notice that distillation often requires task boundaries to know when to update the teacher.

\paragraph{Classifier.} Most methods use a linear classifier trained by backpropagation. Another popular choice is the NCM (Nearest-Class-Mean) classifier, which computes a prototype for each class and uses the distance between prototypes to classify at inference time. For example, SCR uses an NCM classifier. Usually, the NCM classifier is only used during inference, while a separate linear classifier is trained via backpropagation during training.

\paragraph{Model update.} Most methods use end-to-end backpropagation using both the new and the memory samples. Instead, A-GEM \cite{chaudhry2018efficient} uses the gradient from the memory to impose constraints on the update of the model, exploiting the fact that interference can be measured using the cosine similarity between gradients of different tasks.

\paragraph{Baseline methods.} GDumb updates the memory via class-balanced reservoir sampling \cite{chrysakis2020online} and, before each inference step, retrains the entire model using only the memory data. GDumb is not an effective online method because it cannot do anytime inference due to the high cost of retraining at each step. However, it is a useful baseline that is surprisingly effective.


\section{Metrics}
\label{sec:metrics}

To evaluate the performance of each strategy, we use a comprehensive set of metrics recently introduced in the OCL literature. We have taken special care to include metrics to measure stability and knowledge accumulation. Following \cite{lange2023continual}, we perform continual evaluation after every timestep $t$ from the stream (after training on each mini-batch). In addition, we also evaluate the performance at task boundaries. We indicate the accuracy of a model $f$ at training iteration $t$ on evaluation task $E_i$ with $A(E_i, f_t)$, we denote $k$ the current training task index. 

\paragraph{Stability.} The \emph{Worst-Case Accuracy} (WC-ACC) \cite{lange2023continual} provides a trade-off between the accuracy on iteration $t$ and the average minimum accuracy over all tasks learned before the current task $T_k$. Formally:
\begin{equation}
    \text{WC-ACC}_t = \frac{1}{k} A(E_k, f_t) + \bigg(1 - \frac{1}{k}\bigg) \text{min-ACC}_{T_k}.
\end{equation}
The metric \emph{min-ACC} is defined in \cite{lange2023continual} with respect to the last task $T_k$ and can be computed as:
\begin{equation}
    \text{min-ACC}_{T_k} = \frac{1}{k-1} \sum_{i=1}^{k-1} \min_{|T_k|<n\leq t} A(E_i, f_n),
\end{equation}
where $|T_k|$ represents the last training iteration for task $T_k$.

\paragraph{Average Anytime Accuracy.} The \emph{Average Anytime Accuracy} \cite{caccia2021new} ($AAA$), sometimes called Area Under the Curve accuracy \cite{koh2022online} ($A_{AUC}$), is a generalization of the average incremental accuracy \cite{rebuffi2017icarl} to the continual evaluation setting, and is defined as:
\begin{equation}
    AAA_t = \frac{1}{t}\sum_{j=1}^t \frac{1}{k} \sum_{i=1}^k A(E_i, f_j),
\end{equation}

\paragraph{Average Accuracy/Forgetting.} Similarly to most of the works in CL, we report the \emph{Average Accuracy} and the \emph{Average Forgetting} as defined by \cite{lopez2017gradient}. The Average Accuracy is computed on $k$ tasks from a model at step $t$ by $AA = \frac{1}{k} \sum_{i=1}^k A(E_i, f^t)$. Similarly, the Average Forgetting is computed by $AF = \frac{1}{k} \sum_{i=1}^k A(E_i, f^{|T_i|}) - A(E_i, f^t)$.

\paragraph{Cumulative Accuracy/Forgetting.} When applied to class-incremental learning, as done in this paper, the above-described Average Forgetting metric measures both the forgetting and the drop in performance because of the increasingly difficult classification task. Soutif et al.~\cite{soutif2021importance} propose a forgetting measure tailored to class-incremental learning called cumulative forgetting. The \emph{Cumulative Accuracy} for a model $f^t$ is computed on the concatenation of all evaluation tasks seen up to task $k$ ($E_\Sigma^k$), and only considering logits up to the classes seen in task $k$ ($C_\Sigma^k$). It can be computed as
\begin{equation}
    b_k^t = \frac{1}{|E_\Sigma^k|} \sum_{x, y \in E_\Sigma^k} 1_{y}(\argmax_{c \in C_\Sigma^k} f^t(x)_c),
\end{equation}
where $1_{y}(\hat{y})$ is the indicator function that is $1$ if $y=\hat{y}$ and $0$ otherwise. We then compute the \emph{Average Cumulative Forgetting} \cite{soutif2021importance} across all tasks, which is simply computed from the Cumulative Accuracy as $F_\Sigma^t = \frac{1}{t-1} \sum_{k=1}^{t-1} \max_{i=1,\ldots,t}\ b_k^i - b_k^t$.

\paragraph{Representation quality.} In this setup, the model backbone is frozen and a linear classifier is trained on top of it using all the training data seen so far (linear probing)~\cite{davari2022probing}, we then report the accuracy obtained with this classifier (Probed Accuracy). This metric allows us to evaluate the quality of the representations computed with the incrementally learned backbone. This metric is computed only at task boundaries. 

\section{Experimental setup}
\label{sec:setup}

\paragraph{Benchmarks.} We present results on two continual learning classification benchmarks. \textbf{Split-Cifar100} is created from the Cifar-100 dataset \cite{cifar}, that contains 50 000 training images and 10 000 test images of size 32x32, equally divided into 100 classes that are refined from 20 super-classes. \textbf{Split-TinyImagenet} \cite{tinyImgNet2018} is a reduced version of the Imagenet dataset \cite{deng2009imagenet}, containing 100 000 training images resized to 64x64 and splitted in 200 classes. We split these datasets into 20 tasks using a random class order (task composition change for each random seed).

\paragraph{Model and training details.} In all the experiments, we use a slim version of Resnet18 as done in previous works \cite{lopez2017gradient}. We use the SGD optimizer without momentum nor weight decay. We tune the learning rate for each method using hyperparameter selection protocol defined later in this section. Since all of the compared methods make use of a replay buffer, we choose to follow a common protocol for learning that consists in performing several training passes on the same batch of data, using a different batch of memory. This protocol has been used in several works \cite{MIR, zhang2022simple, hu2021onepassimagenet}. We choose to always apply input transformations since they have been proven to drastically reduce the overfitting on the buffer samples and to be efficient in combination with several iterations per incoming batch \cite{zhang2022simple}. We use a batch size of 10 for the current data for both benchmarks. Each training batch is then constituted of 10 images from current data and 10 images from the replay buffer, except for SCR which needs to sample a bigger batch from memory (of size 118). The number of training passes on each mini-batch is kept fixed for all methods (not tuned), and set to 3 on Split-Cifar100 and 9 on Split-TinyImagenet. On both of the benchmarks we use random cropping and random horizontal flip as input transformation, except for SCR which uses more advanced transformations. In the main results, we use a fixed memory size of 2000 on Split-Cifar100 and 4000 on Split-TinyImagenet. Additionally, we present results for more extreme amounts of memory (low and high) on Split-Cifar100 in Figure \ref{fig:cifar_analysis}.

\paragraph{Hyperparameter selection.} In order to make sure every method tried performs at its best, we use a hyperparameter selection mechanism. The constraints of the online setting usually do not allow for efficient hyperparameter selection. In \cite{mai2022online}, the first few tasks of the training stream are used as validation tasks, which is borrowed from the protocol defined in \cite{chaudhry2018efficient}. We also follow this protocol, using the first 4 tasks (out of 20), to optimize the hyperparameters by looking at the accuracy on the validation set. Contrary to what is done in \cite{chaudhry2018efficient}, we do not allow the learner to learn offline on these 4 tasks, but rather put the learner in the setting of online learning. We use the tree-structured Parzen estimator algorithm~\cite{bergstra2011algorithms} to guide the search of hyperparameters, by running 200 trials for each method. In the Appendix (See Section. \ref{sec:hpselection}) we provide the list of tuned parameters for each method as well as the ranges used and the best values found.

\paragraph{Evaluation.} As done in \cite{caccia2021new, koh2022online}, and studied more in depth in \cite{de2021continual}. We perform \textit{continual evaluation}, meaning we evaluate the model on held-out validation data (5\% of the whole stream) after learning on each new mini-batch. On top of this, we also perform a more classical evaluation on the test stream after training on each task.

\paragraph{Methods.} We report results for all methods presented in Section \ref{sec:methods}. On top of that, we add an i.i.d. reference method, which uses the same methodology as the one of ER but learns on an i.i.d. stream instead of a class-incremental one.

\paragraph{Implementation.} All methods were implemented using the Avalanche framework~\cite{carta2023avalanche}, an open source continual learning framework that provides the tools to implement new strategies and benchmarks in continual learning. While some methods were already present, we adapted some of the existing methods and added some new methods to the framework to conduct this study. The implementation of each method is modular, which allows reuse of common components (e.g., reservoir buffers) and highlights the similarities and differences between methods. Despite our efforts to make the comparison as fair as possible, there are some implementation choices that made it difficult to reconcile all methods. We list them in the Appendix (See Section. \ref{sec:implementation}) and discuss their potential impact on the results.

\section{Results}\label{sec:results}

\tablemerged{Last step results on Split-Cifar100 (20 Tasks) with 2000 memory (Left) and for Split-TinyImagenet (20 Tasks) with 4000 memory (Right). For each metric, we report the average and standard deviation over 5 seeds}

\paragraph{Final Average Accuracy.} On \textbf{Split-Cifar100}, we found that the final performance of all compared methods is very similar to the one of vanilla replay (ER) (See Table \ref{tab:accuracies}). Except for the performance of AGEM, most methods performed quite competitively in the OCL setting (only around 5\% away from the i.i.d. reference method), the best one being the introduced baseline combining ER and LwF, but only by a tiny margin. On \textbf{Split-TinyImagenet}, the results are similarly close (See Table \ref{tab:accuracies}), with the exception of ER-ACE performing better on that benchmark, and RAR and SCR underperforming.

\paragraph{Stability ($\operatorname{WC-Acc}$ and $AAA$).} On \textbf{Split-Cifar100}, in terms of stability, the results vary much more between each method, and the final performance is not necessarily correlated with the stability of the method. For instance, the final accuracy for MIR is $1\%$ above the final accuracy for SCR, however, SCR has about $9\%$ better $\operatorname{WC-Acc}$ than MIR. In Figure \ref{fig:cifar_analysis}, we illustrate the difference in stability between SCR and ER on Split-Cifar100 with 2000 memory. In this figure, the accuracy on the previous task drops significantly when shifting tasks in the case of ER, indicating low stability, this is not the case for SCR. In general, we observe that the $AAA$ is moderately correlated to the $\operatorname{WC-Acc}$, even though they are not strongly linked in theory (it is possible to have low $\operatorname{WC-Acc}$ and high $AAA$).

\paragraph{Representation quality.} Surprisingly, we find that the probed accuracy for most methods is close to the one of the i.i.d. reference method ($45.8\%$ on Split-Cifar100 and $34.3\%$ on Split-Tinyimagenet). 
This suggests that the representation learned by methods on the continual stream is not much worse than the one learned on the i.i.d. stream. Therefore, the significant performance difference (in \emph{Acc}) between incremental learning methods and the i.i.d. reference is most likely caused by a deterioration of the incrementally learned classifier. In the Appendix (See Tables \ref{tab:cifar500} and \ref{tab:cifar8k}), we provide results with 500 and 8000 memory on Split-Cifar100. Even when using 500 memory, we observe a similar conclusion, with the gap to probing augmenting only a bit (from $1\%$ to $2\%$), showing that only a low amount of memory (5 per class in that case) is efficient to get a decent representation strength. In general, we find that the probed accuracy is not strongly linked with the stability metrics. In Table \ref{tab:accuracies}, we see that ER has in both cases the best Probed Accuracy, while it has lower than average stability metrics compared to the other methods.

\paragraph{Forgetting.} In Figure \ref{fig:cifar_analysis}, we display both the classical forgetting and the cumulative forgetting defined in Section \ref{sec:metrics} \cite{soutif2021importance}. We see that while the classical forgetting indicates a high amount of forgetting that is increasing across the stream, the cumulative forgetting gives a different picture, indicating that for all methods, some backward transfer is achieved, and this remains quite constant across all tasks. Two curves exhibit slightly distinct behavior, namely the ones of SCR and MIR. These distinctive behaviors can also be observed in Figure~\ref{fig:comparison}. In the case of MIR, the average accuracy is initially low, and later increases to match the one of $\operatorname{ER-ACE}$, resulting in high backward transfer. Whereas for SCR, the accuracy is initially high, but later meets the one of other methods, resulting in neutral forgetting (no forgetting, but no backward transfer). Classical forgetting however is not very relevant in the class-incremental learning setting because it increases consequently to an increase in the difficulty of the task (more and more classes to consider in the classification problem), which makes it hard to interpret, as such, we advise against its use in class-incremental learning (online or not online). In definitive, these forgetting numbers indicate that there is some backward transfer happening, we believe this is mainly due to the fact that the network is underfitted in online learning, due to the low number of training iterations, making it easy to gain additional performance on a task when training on subsequent tasks.

\paragraph{Effect of the memory batch size.} As explained in Section \ref{sec:setup}, the memory batch size is set to the same as the current batch  size in our experiments, except for SCR, which requires a higher one. We believe that this difference explains the good performance of SCR in the early training regime (see Figure \ref{fig:comparison}) and in the high memory size setting. We perform additional experiments (See Table \ref{tab:cifar8k}) where we use the same memory batch size for ER as the one of SCR. On Split-Cifar100 with 8000 memory, changing the memory batch size from 10 to 118 is sufficient to make ER match the performance of SCR (jumping from $34.9\%$ to $43.0\%$ final accuracy). This confirms our belief that this parameter is important to take into account when interpreting results.

\paragraph{Effect of the memory size.} In Figure \ref{fig:cifar_analysis}, we report the final performance of ER, the i.i.d. reference method, and the GDumb baseline when using more extreme (low and high) memory amounts on Split-Cifar100. When high amounts of memory are used, the GDumb baseline can surpass the performance of the continual learning methods if no special care is given to the memory batch size. This one needs to be adapted in order to obtain the best performances with continual learning methods (see Appendix Section. \ref{sec:additionalresults}).

\figcomparison{20 100 20 20}{1.0}{Validation stream accuracy for each of the methods, compared to the one of the i.i.d. reference method, on Split-Cifar100, using 2000 exemplars (Left), and Split-TinyImagenet, using 4000 exemplars (Right). The accuracy is reported after training on each mini-batch, we display mean and standard deviation across 5 seeds.}






\figcifarother{0 0 0 0}{1.0}{\textbf{Left:} Forgetting (full lines), and Cumulative Forgetting (dotted lines) on Split-Cifar100 with 2000 memory; \textbf{Middle:} Illustration of the difference in stability between ER and SCR on Split-Cifar100 (20 tasks), using 2000 memory. We place ourselves at the task shift between task 4 and 5 and display the accuracy on previous task data (dotted lines) as well as the accuracy on current task data (full lines).; \textbf{Right:} Final performance of ER, i.i.d. reference method, and GDumb baseline for 3 different memory sizes on Split-Cifar100}

\section{Related Work}

\paragraph{Existing surveys.} Surveys on continual learning have focused on different aspects. Parisi et al.~\cite{parisi2019continual} provide a survey on lifelong learning and draw parallels with how biological systems prevent catastrophic forgetting. The survey of Lesort et al.~\cite{lesort2020continual} studies continual learning focusing on robotics applications. More recent surveys have focused on popular settings for continual learning. De Lange et al.~\cite{de2021continual} studies task-incremental learning, and Masana et al.~\cite{masana2022class} investigates class-incremental learning. More similar to the proposed work in this paper, is the work of Mai et al.~\cite{mai2022online} who propose an empirical evaluation of several online continual learning methods. However, other than them, our paper aims to compare a variety of competitive replay methods that each use different approaches to tackle the setting of online continual learning (as described in Figure \ref{fig:ocl-algo}). On top of that, we evaluate and compare the performance of each method from different points of view, analyzing both the final performance but also the stability (by evaluating on the validation set after each mini-batch). We also evaluate the learned representation by probing the features with the full dataset at the end of the training, as in \cite{davari2022probing}.



\paragraph{Existing Libraries.} This survey contributes to the implementation of multiple methods in the Avalanche Library \cite{carta2023avalanche}. Avalanche is an end-to-end library based on Pytorch with the goal of providing a codebase for fast prototyping, training, and reproducible evaluation of continual learning algorithms. Besides this, other libraries have been implemented with different objectives and qualities. SequeL \cite{dimitriadis2023sequel} is a new library that focuses on developing methods not only in PyTorch, but also in JAX. It provides a simple interface for running experiments in both. However, the newness of the library and the need to implement the methods in both languages make it difficult to use. On the other hand, Sequoia \cite{normandin2021sequoia} is a library that attempts to unite as many continual learning settings as possible under a common tree. The root is the most difficult problem to learn, and the leaves and branches are different settings. Lastly, Continuum \cite{douillardlesort2021continuum} is a library that focuses mostly on the benchmark aspects of continual learning and provides tools to easily split the datasets and iterate on the resulting tasks.


\paragraph{Additional methods.} In addition to the methods implemented, there are other methods proposed in recent years. In Online Bias Correction \cite{chrysakis2022online}, the authors explain how experience replay biases the model output towards recent observations. With this, they propose a way to modify the classifier output and mitigate the bias. Following the same idea of reducing the bias, Guo et al. \cite{guo2022online} propose OCM based on mutual information maximization. Here, the authors deal with the bias reduction caused by cross entropy and they encourage the preservation of previous knowledge.
Another approach that relies on a buffer is Proxy-based Contrastive Replay \cite{lin2023pcr}. Here, the authors propose a way to complement a buffer loss and a contrastive loss. Using a Visual Transformer in conjunction with a focal contrastive learning strategy, Wang et al. \cite{wang2022online} suggest mitigating the stability-plasticity dilemma.





\section{Conclusions}

\paragraph{} In this study, we examined the performance of various Online Continual Learning (OCL) methods, focusing on performance, stability, representation quality, and forgetting. Our analysis revealed intriguing insights. Firstly, we found that stability does not always translate to higher accuracy, challenging the notion that a stable model guarantees superior performance in the OCL setting. Additionally, we observed that the quality of the representation learned by continual learning methods does not differ strongly from the one obtained by learning on the i.i.d. stream, indicating that the main challenge faced by continual learning methods is to learn a good classifier. We also found that methods were prone to underfitting in the OCL setting, challenging the common assumption that continual learning methods suffer from forgetting; we here claim that they keep improving their performance on previous tasks as they learn on subsequent tasks. In general, we found all compared methods to perform very similarly to the common Experience Replay (ER)  method. We also investigate some small implementation differences and conclude that sometimes small details in implementations can make a method shine using the existing metrics, but that it is often possible to obtain these same results by slightly modifying the baseline ER hyperparameters or implementation details, highlighting the necessity to implement these methods in a unified framework like avalanche so that they can be more fairly compared. Finally, we found that no single OCL method proved to be universally superior across all metrics or memory sizes, highlighting the absence of a one-fits-all solution. Considering these findings, we identify several promising research directions for online continual learning.

\paragraph{} We advocate for stronger connections between normal i.i.d. online learning and online continual learning, given their similar representation strengths. As the ER strategy is already common to both settings and proven competitive, emphasis should be put on tuning its hyperparameters during training, as already attempted in \cite{zhang2022closedloop} and \cite{cai2021online}. Proper hyperparameter tuning in online continual learning remains an open challenge. Additionally, we encourage further exploration of linking stability metrics to training efficiency, as we found that poor stability does not necessarily impact final representation strength. If no direct link exists, enforcing good stability during training may not be essential, and ad-hoc methods~\cite{soutifcormerais2023improving} could be sufficient to achieve desired stability.

\paragraph{Acknowledgements:} We acknowledge the support from the Spanish Government funding for projects PID2022-143257NB-I00, TED2021-132513B-I00. We also acknowledge the support from the Italian Ministry of University and Research (MUR) as part of the FSE REACT-EU - PON 2014-2020 “Research and Innovation" resources – Innovation Action - DM MUR 1062/2021

\bibliography{egbib,antonio}

\begin{thebibliography}{10}\itemsep=-1pt

\bibitem{tinyImgNet2018}
Stanford, “tiny imagenet challenge, cs231n course.”.
\newblock \url{https://tiny-imagenet.herokuapp.com/}, 2015.

\bibitem{MIR}
Rahaf Aljundi, Eugene Belilovsky, Tinne Tuytelaars, Laurent Charlin, Massimo
  Caccia, Min Lin, and Lucas Page-Caccia.
\newblock Online continual learning with maximal interfered retrieval.
\newblock In H. Wallach, H. Larochelle, A. Beygelzimer, F. d\textquotesingle
  Alch\'{e}-Buc, E. Fox, and R. Garnett, editors, {\em Advances in Neural
  Information Processing Systems}, volume~32. Curran Associates, Inc., 2019.

\bibitem{aljundi2019gradient}
Rahaf Aljundi, Min Lin, Baptiste Goujaud, and Yoshua Bengio.
\newblock Gradient based sample selection for online continual learning.
\newblock {\em Advances in neural information processing systems}, 32, 2019.

\bibitem{bergstra2011algorithms}
James Bergstra, R{\'e}mi Bardenet, Yoshua Bengio, and Bal{\'a}zs K{\'e}gl.
\newblock Algorithms for hyper-parameter optimization.
\newblock {\em Advances in neural information processing systems}, 24, 2011.

\bibitem{buzzega2020dark}
Pietro Buzzega, Matteo Boschini, Angelo Porrello, Davide Abati, and Simone
  Calderara.
\newblock Dark experience for general continual learning: a strong, simple
  baseline.
\newblock {\em Advances in neural information processing systems},
  33:15920--15930, 2020.

\bibitem{caccia2021new}
Lucas Caccia, Rahaf Aljundi, Nader Asadi, Tinne Tuytelaars, Joelle Pineau, and
  Eugene Belilovsky.
\newblock New insights on reducing abrupt representation change in online
  continual learning.
\newblock In {\em International Conference on Learning Representations}, 2021.

\bibitem{cai2021online}
Zhipeng Cai, Ozan Sener, and Vladlen Koltun.
\newblock Online continual learning with natural distribution shifts: An
  empirical study with visual data.
\newblock In {\em Proceedings of the IEEE/CVF international conference on
  computer vision}, pages 8281--8290, 2021.

\bibitem{carta2023avalanche}
Antonio Carta, Lorenzo Pellegrini, Andrea Cossu, Hamed Hemati, and Vincenzo
  Lomonaco.
\newblock Avalanche: A pytorch library for deep continual learning.
\newblock {\em arXiv preprint arXiv:2302.01766}, 2023.

\bibitem{chaudhry2018efficient}
Arslan Chaudhry, Marc’Aurelio Ranzato, Marcus Rohrbach, and Mohamed
  Elhoseiny.
\newblock Efficient lifelong learning with a-gem.
\newblock In {\em International Conference on Learning Representations}, 2018.

\bibitem{chaudhry2019tiny}
Arslan Chaudhry, Marcus Rohrbach, Mohamed Elhoseiny, Thalaiyasingam Ajanthan,
  Puneet~K Dokania, Philip~HS Torr, and Marc'Aurelio Ranzato.
\newblock On tiny episodic memories in continual learning.
\newblock {\em arXiv preprint arXiv:1902.10486}, 2019.

\bibitem{chrysakis2020online}
Aristotelis Chrysakis and Marie-Francine Moens.
\newblock Online continual learning from imbalanced data.
\newblock In {\em International Conference on Machine Learning}, pages
  1952--1961. PMLR, 2020.

\bibitem{chrysakis2022online}
Aristotelis Chrysakis and Marie-Francine Moens.
\newblock Online bias correction for task-free continual learning.
\newblock In {\em The Eleventh International Conference on Learning
  Representations}, 2022.

\bibitem{cossuContinualPreTrainingMitigates2022}
Andrea Cossu, Tinne Tuytelaars, Antonio Carta, Lucia Passaro, Vincenzo
  Lomonaco, and Davide Bacciu.
\newblock Continual {{Pre-Training Mitigates Forgetting}} in {{Language}} and
  {{Vision}}, May 2022.

\bibitem{davari2022probing}
MohammadReza Davari, Nader Asadi, Sudhir Mudur, Rahaf Aljundi, and Eugene
  Belilovsky.
\newblock Probing representation forgetting in supervised and unsupervised
  continual learning.
\newblock In {\em Proceedings of the IEEE/CVF Conference on Computer Vision and
  Pattern Recognition}, pages 16712--16721, 2022.

\bibitem{de2021continual}
Matthias De~Lange, Rahaf Aljundi, Marc Masana, Sarah Parisot, Xu Jia,
  Ale{\v{s}} Leonardis, Gregory Slabaugh, and Tinne Tuytelaars.
\newblock A continual learning survey: Defying forgetting in classification
  tasks.
\newblock {\em IEEE transactions on pattern analysis and machine intelligence},
  44(7):3366--3385, 2021.

\bibitem{deng2009imagenet}
Jia Deng, Wei Dong, Richard Socher, Li-Jia Li, Kai Li, and Li Fei-Fei.
\newblock Imagenet: A large-scale hierarchical image database.
\newblock In {\em 2009 IEEE conference on computer vision and pattern
  recognition}, pages 248--255. Ieee, 2009.

\bibitem{dimitriadis2023sequel}
Nikolaos Dimitriadis, Francois Fleuret, and Pascal Frossard.
\newblock Sequel: A continual learning library in pytorch and jax.
\newblock {\em arXiv preprint arXiv:2304.10857}, 2023.

\bibitem{douillardlesort2021continuum}
Arthur Douillard and Timothée Lesort.
\newblock Continuum: Simple management of complex continual learning scenarios,
  2021.

\bibitem{finiSelfSupervisedModelsAre2022}
Enrico Fini, Victor G.~Turrisi Da~Costa, Xavier {Alameda-Pineda}, Elisa Ricci,
  Karteek Alahari, and Julien Mairal.
\newblock Self-{{Supervised Models}} are {{Continual Learners}}.
\newblock In {\em 2022 {{IEEE}}/{{CVF Conference}} on {{Computer Vision}} and
  {{Pattern Recognition}} ({{CVPR}})}, pages 9611--9620, June 2022.

\bibitem{guo2022online}
Yiduo Guo, Bing Liu, and Dongyan Zhao.
\newblock Online continual learning through mutual information maximization.
\newblock In {\em International Conference on Machine Learning}, pages
  8109--8126. PMLR, 2022.

\bibitem{hastie2009elements}
Trevor Hastie, Robert Tibshirani, Jerome~H Friedman, and Jerome~H Friedman.
\newblock {\em The elements of statistical learning: data mining, inference,
  and prediction}, volume~2.
\newblock Springer, 2009.

\bibitem{hu2021onepassimagenet}
Huiyi Hu, Ang Li, Daniele Calandriello, and Dilan G{\"{o}}r{\"{u}}r.
\newblock One pass imagenet.
\newblock {\em CoRR}, abs/2111.01956, 2021.

\bibitem{koh2022online}
Hyunseo Koh, Dahyun Kim, Jung-Woo Ha, and Jonghyun Choi.
\newblock Online continual learning on class incremental blurry task
  configuration with anytime inference.
\newblock In {\em International Conference on Learning Representations}, 2022.

\bibitem{cifar}
Alex Krizhevsky.
\newblock Learning multiple layers of features from tiny images.
\newblock Technical report, 2009.

\bibitem{kumari2022retrospective}
Lilly Kumari, Shengjie Wang, Tianyi Zhou, and Jeff~A Bilmes.
\newblock Retrospective adversarial replay for continual learning.
\newblock {\em Advances in Neural Information Processing Systems},
  35:28530--28544, 2022.

\bibitem{lange2023continual}
Matthias~De Lange, Gido~M van~de Ven, and Tinne Tuytelaars.
\newblock Continual evaluation for lifelong learning: Identifying the stability
  gap.
\newblock In {\em The Eleventh International Conference on Learning
  Representations}, 2023.

\bibitem{lesort2020continual}
Timoth{\'e}e Lesort, Vincenzo Lomonaco, Andrei Stoian, Davide Maltoni, David
  Filliat, and Natalia D{\'\i}az-Rodr{\'\i}guez.
\newblock Continual learning for robotics: Definition, framework, learning
  strategies, opportunities and challenges.
\newblock {\em Information fusion}, 58:52--68, 2020.

\bibitem{li2017learning}
Zhizhong Li and Derek Hoiem.
\newblock Learning without forgetting.
\newblock {\em IEEE transactions on pattern analysis and machine intelligence},
  40(12):2935--2947, 2017.

\bibitem{lin2023pcr}
Huiwei Lin, Baoquan Zhang, Shanshan Feng, Xutao Li, and Yunming Ye.
\newblock Pcr: Proxy-based contrastive replay for online class-incremental
  continual learning.
\newblock In {\em Proceedings of the IEEE/CVF Conference on Computer Vision and
  Pattern Recognition}, pages 24246--24255, 2023.

\bibitem{lopez2017gradient}
David Lopez-Paz and Marc'Aurelio Ranzato.
\newblock Gradient episodic memory for continual learning.
\newblock {\em Advances in neural information processing systems}, 30, 2017.

\bibitem{madaanRepresentationalContinuityUnsupervised2022}
Divyam Madaan, Jaehong Yoon, Yuanchun Li, Yunxin Liu, and Sung~Ju Hwang.
\newblock Representational {{Continuity}} for {{Unsupervised Continual
  Learning}}.
\newblock In {\em {{ICLR}}}, Apr. 2022.

\bibitem{mai2022online}
Zheda Mai, Ruiwen Li, Jihwan Jeong, David Quispe, Hyunwoo Kim, and Scott
  Sanner.
\newblock Online continual learning in image classification: An empirical
  survey.
\newblock {\em Neurocomputing}, 469:28--51, 2022.

\bibitem{mai2021supervised}
Zheda Mai, Ruiwen Li, Hyunwoo Kim, and Scott Sanner.
\newblock Supervised contrastive replay: Revisiting the nearest class mean
  classifier in online class-incremental continual learning.
\newblock In {\em Proceedings of the IEEE/CVF Conference on Computer Vision and
  Pattern Recognition}, pages 3589--3599, 2021.

\bibitem{masana2022class}
Marc Masana, Xialei Liu, Bart{\l}omiej Twardowski, Mikel Menta, Andrew~D
  Bagdanov, and Joost Van De~Weijer.
\newblock Class-incremental learning: survey and performance evaluation on
  image classification.
\newblock {\em IEEE Transactions on Pattern Analysis and Machine Intelligence},
  45(5):5513--5533, 2022.

\bibitem{mohri2018foundations}
Mehryar Mohri, Afshin Rostamizadeh, and Ameet Talwalkar.
\newblock {\em Foundations of machine learning}.
\newblock MIT press, 2018.

\bibitem{normandin2021sequoia}
Fabrice Normandin, Florian Golemo, Oleksiy Ostapenko, Pau Rodriguez, Matthew~D
  Riemer, Julio Hurtado, Khimya Khetarpal, Ryan Lindeborg, Lucas Cecchi,
  Timoth{\'e}e Lesort, et~al.
\newblock Sequoia: A software framework to unify continual learning research.
\newblock {\em arXiv preprint arXiv:2108.01005}, 2021.

\bibitem{parisi2019continual}
German~I Parisi, Ronald Kemker, Jose~L Part, Christopher Kanan, and Stefan
  Wermter.
\newblock Continual lifelong learning with neural networks: A review.
\newblock {\em Neural networks}, 113:54--71, 2019.

\bibitem{prabhu2020gdumb}
Ameya Prabhu, Philip~HS Torr, and Puneet~K Dokania.
\newblock Gdumb: A simple approach that questions our progress in continual
  learning.
\newblock In {\em Computer Vision--ECCV 2020: 16th European Conference,
  Glasgow, UK, August 23--28, 2020, Proceedings, Part II 16}, pages 524--540.
  Springer, 2020.

\bibitem{rebuffi2017icarl}
Sylvestre-Alvise Rebuffi, Alexander Kolesnikov, Georg Sperl, and Christoph~H
  Lampert.
\newblock icarl: Incremental classifier and representation learning.
\newblock In {\em Proceedings of the IEEE conference on Computer Vision and
  Pattern Recognition}, pages 2001--2010, 2017.

\bibitem{soutifcormerais2023improving}
Albin Soutif-Cormerais, Antonio Carta, and Joost~Van de Weijer.
\newblock Improving online continual learning performance and stability with
  temporal ensembles, 2023.

\bibitem{soutif2021importance}
Albin Soutif-Cormerais, Marc Masana, Joost Van~de Weijer, and Bartl{\o}miej
  Twardowski.
\newblock On the importance of cross-task features for class-incremental
  learning.
\newblock {\em arXiv: 2106.11930}, 2021.

\bibitem{wang2022online}
Zhen Wang, Liu Liu, Yajing Kong, Jiaxian Guo, and Dacheng Tao.
\newblock Online continual learning with contrastive vision transformer.
\newblock In {\em European Conference on Computer Vision}, pages 631--650.
  Springer, 2022.

\bibitem{zhang2022closedloop}
Yaqian Zhang, Eibe Frank, Bernhard Pfahringer, Albert Bifet, Nick Jin~Sean Lim,
  and Alvin Jia.
\newblock Closed-loop control for online continual learning, 2022.

\bibitem{zhang2022simple}
Yaqian Zhang, Bernhard Pfahringer, Eibe Frank, Albert Bifet, Nick Jin~Sean Lim,
  and Yunzhe Jia.
\newblock A simple but strong baseline for online continual learning: Repeated
  augmented rehearsal.
\newblock {\em Advances in Neural Information Processing Systems},
  35:14771--14783, 2022.

\end{thebibliography}
\bibliographystyle{ieee_fullname}

\section{Appendix}

\subsection{Additional results}

\label{sec:additionalresults}

In Table \ref{tab:cifar500} and \ref{tab:cifar8k}, we present results for more extreme amounts of memory (lower and higher) on Split-Cifar100. For the low memory setting, we notice that the final accuracy results differ more between each method than when using 2000 memory, with ER-ACE getting the best results both in terms of final accuracy and stability. However, the probed accuracy is still close to the one of the i.i.d reference method. When using more memory, we see that the performance of the GDumb baselines matches the one of the i.i.d reference method. However, SCR surpasses both of these, indicating that it's still possible to learn more from the whole stream than from just the memory. We suppose that this is due to its use of a bigger memory batch size, which would be beneficial when a bigger memory size is used. To verify this, we perform an additional experiment where we provide ER with the same memory batch size as SCR, we see that with this modification, the performance of ER matches the one of SCR, indicating that the performance of SCR in this setting is probably due to the bigger memory batch size and not so much to the supervised-contrastive loss.  

\tablecifarsmall{Last step results on Split-Cifar100 (20 Tasks) with 500 memory. We report the average and standard deviation over 5 trials}

\tablecifarbig{Last step results on Split-Cifar100 (20 Tasks) with 8000 memory. We report the average and standard deviation over 5 trials}

\subsection{Implementation Details}

\label{sec:implementation}

Despite our efforts to make the comparison as fair as possible, there are a few points on which it was hard to make every method coincide. We list them in the following section:

\begin{itemize}
    \item \textbf{Handling of batch normalisation statistics}: While sampling a batch from the current task and the memory, there is a choice that needs to be made when forwarding each batch to the model. The default solution adopted in Avalanche is to concatenate both batches and perform one pass on the model using the concatenated batch statistics (option 1). However, some methods were initially implemented by forwarding each batch separately, which could have a huge influence since in that case the separate outputs are created using each internal batch statistics (option 2). In general, while implementing the methods, we chose the option that was working best (ER: 1, DER++: 1, ER-ACE: 2, MIR: 2, SCR: 1, RAR: 2). Note that MIR also updates the batchnorm statistics when forwarding the bigger replay batch (from which it selects the samples to replay), which also has an influence on training that other methods do not have.
    \item \textbf{Memory batch size}: Initially, we wanted to fix the batch size memory using the hyperparameter validation protocol described above, so that each method could select it's adequate memory batch size. However, we found that when using a fixed memory size and doing the hyperparameter selection on only 4 tasks, a big memory batch size was always selected since it was giving more beneficial results after seeing only 4 tasks. This is due to the fact that the optimal use of the full memory size is close to always iterating on samples from the memory. Because of this, we chose to also fix the memory batch size to the same size as the one of the current batch (as done in most works). However, due to its use of a contrastive loss, SCR requires to sample a big batch from the memory, so we fixed the memory batch size to a higher number (118), which makes it behave differently than other methods.
    \item \textbf{Dynamic Classifier}: In continual learning, the learner is not suppose to know the total number of classes it will encounter during the training. This is why we implemented most methods using a dynamic classification layer that adds new units whenever encountering a new class. However, one method (DER++) requires to replay the logits of samples from previous classes into the new classes. The official implementation made use of a classification layer of fixed size, and we used the same in our experiments, making it different from what other methods do.
\end{itemize}

\subsection{Hyperparameters}

\label{sec:hpselection}

In the code (\url{https://github.com/AlbinSou/ocl_survey}), we provide the script used to perform the hyperparameter selection (experiments/main\_hp\_tuning.py) as well as the best configurations we found after 200 trials for each method using the first 4 tasks of the stream. We provide these configurations for each benchmark under the config/best\_configs folder. The hyperparameter ranges tested for each method are available in experiments/spaces.py.

\end{document}